\documentclass[pdflatex]{sn-jnl}

\usepackage{graphicx}%
\usepackage{multirow}%
\usepackage{amsmath,amssymb,amsfonts}%
\usepackage{amsthm}%
\usepackage{mathrsfs}%
\usepackage[title]{appendix}%
\usepackage{xcolor}%
\usepackage{textcomp}%
\usepackage{manyfoot}%
\usepackage{booktabs}%
\usepackage{algorithm}%
\usepackage{algorithmicx}%
\usepackage{algpseudocode}%
\usepackage{listings}%

\begin{document}

\title[Big Data - Supply Chain Management Framework for Forecasting: Data Preprocessing and Machine Learning Techniques]{Big Data -- Supply Chain Management Framework for Forecasting: Data Preprocessing and Machine Learning Techniques}


\author[1]{\fnm{Md Abrar} \sur{Jahin}}\email{abrar.jahin.2652@gmail.com}

\author[2]{\fnm{Md Sakib Hossain} \sur{Shovon}}\email{sakib.aiub.cs@gmail.com}

\author [3]{\fnm{Jungpil} \sur{Shin}}\email{jpshin@u-aizu.ac.jp}

\author[4]{\fnm{Istiyaque Ahmed} \sur{Ridoy}}\email{istiyaque63@gmail.com}

\author*[2]{\fnm{M. F.} \sur{Mridha}}\email{firoz.mridha@aiub.edu}

\affil[1]{\orgdiv{Department of Industrial Engineering and Management}, \orgname{Khulna University of Engineering and Technology (KUET)}, \orgaddress{\city{Khulna}, \postcode{9203}, \country{Bangladesh}}}

\affil[2]{\orgdiv{Department of Computer Science}, \orgname{American International University-Bangladesh}, \orgaddress{\city{Dhaka}, \postcode{1229}, \country{Bangladesh}}}

\affil[3]{\orgdiv{Department of Computer Science and Engineering}, \orgname{The University of Aizu}, \orgaddress{\city{Aizuwakamatsu}, \postcode{965-8580}, \country{Japan}}}

\affil[4]{\orgdiv{Institute of Business Administration}, \orgname{University of Dhaka}, \orgaddress{\city{Dhaka}, \country{Bangladesh}}}


\abstract{This article systematically identifies and comparatively analyzes state-of-the-art supply chain (SC) forecasting strategies and technologies within a specific timeframe, encompassing a comprehensive review of 152 papers spanning from 1969 to 2023. A novel framework has been proposed incorporating Big Data Analytics in SC Management (problem identification, data sources, exploratory data analysis, machine-learning model training, hyperparameter tuning, performance evaluation, and optimization), forecasting effects on human workforce, inventory, and overall SC. Initially, the need to collect data according to SC strategy and how to collect them has been discussed. The article discusses the need for different types of forecasting according to the period or SC objective. The SC KPIs and the error-measurement systems have been recommended to optimize the top-performing model. The adverse effects of phantom inventory on forecasting and the dependence of managerial decisions on the SC KPIs for determining model performance parameters and improving operations management, transparency, and planning efficiency have been illustrated. The cyclic connection within the framework introduces preprocessing optimization based on the post-process KPIs, optimizing the overall control process (inventory management, workforce determination, cost, production \& capacity planning). The contribution of this research lies in the standard SC process framework proposal, recommended forecasting data analysis, forecasting effects on SC performance, machine learning algorithms optimization followed, and in shedding light on future research.}




\maketitle

\section{Introduction}
\label{sec:introduction}
In the dynamic landscape of supply chain management (SCM), the relentless pursuit of efficiency and adaptability has driven a continuous evolution in forecasting strategies and technologies. This article embarks on a systematic exploration, aiming to identify and analyze the state-of-the-art in supply chain (SC) forecasting, ultimately proposing a novel framework that integrates the power of big data analytics (BDA) into SCM. The increasing complexity and interconnectedness of global SCs have underscored the need for sophisticated forecasting strategies. Traditional approaches are being reevaluated in the wake of technological advancements, leading to a paradigm shift in how we perceive and optimize SC forecasting. The integration of BDA emerges as a transformative force, promising enhanced predictive capabilities and a holistic framework that spans problem identification, data sourcing, exploratory data analysis, machine learning (ML) model training, hyperparameter tuning, performance evaluation, and optimization.

The SC has evolved sufficiently over the past years to discover new methods and techniques for solving SCM problems. The SC can develop its configuration based on its control, coordination, and management \cite{maccarthy2016}. The advent of big data (BD) brings one such change. Like other fields, BD can be utilized to improve decision-making reprocesses and alter business models through multiple resources, tools, and applications \cite{waller2013}. Therefore, SC and BD usages are connected to help one another. Although the concepts for SCM are already well-developed, it is possible to improve further; recent research on enhancing efficiency through collaboration \cite{schliephake2009}, usage of RFID and intelligent goods \cite{holmqvist2006} are two examples of such innovations that improved SCM processes. Newer technologies are further enabling the discovery of innovative strategies for solving SC problems. BDA is one such disruptive innovation. Although BD has been present for a long time, the approaches to making sense of BD are comparatively new, and such systems have not been wholly integrated into other branches of knowledge \cite{yu2018}. We identified the absence of data usage and relevant processes in SC directly as a major problem that needs to be addressed.

BD has similarly grown popular over the years. After academic and technical publications first mentioned such technological developments, it drew the attention of various people, including literary scholars, corporate leaders, and government officials \cite{hazen2018}. The most recognizable feature of BD is probably its size or the amount of data stored. The distinctive features of greater data variety, high velocity in collection and analysis, the necessity to navigate veracity challenges, and the inherent value growth with increasing data analysis \cite{gandomi2015} set the stage for a new era in SCM. Simply having access to BD is not helpful; data analytics is a must to create value or gather information out of the enormous collection of data. Where analytical methods are applied to BD, it is called BDA. While BDA has vast applications, its role in improving the SC process is notable.

The motivation behind this research stems from the recognition that the traditional forecasting paradigms may not be equipped to address the intricacies of modern SC dynamics. Phantom inventory, varying time horizons, and diverse SCM objectives necessitate a more adaptive and data-driven approach. By introducing a comprehensive framework, this article seeks to bridge the gap between traditional forecasting methods and the demands of contemporary SC environments.

\subsection{Research gaps}
We found potential research gaps combining the preprocessing for ML forecasting, control process (SC processes where BDA is helpful), and post-process (for the evaluation of the forecasting model). Although there is fragmented research on these particular topics (“SC forecasting model performance,” “the application of BDA on SC,” or “ML forecasting techniques with BDA implementation,” or “BD driven SC performance evaluation”), the necessity to form a cyclic connection among these three processes led us to the development of this article. This paper identifies a critical gap in current SCM practices — the underutilization of data and relevant processes - and positions BDA as a powerful solution. The motivation behind this study lies in addressing the pivotal challenge of harnessing the full potential of BD within SCM, recognizing it not just as a technological evolution but as a strategic imperative for future competitiveness.

\subsection{Research objectives}
The primary objective of this article is to shed light on the potential BDA investigations on SCM studies, what significant contributions BDA has made to the efficient use of ML forecasting in SC processes, what preprocessing and post-processing SC forecasting techniques have been robustly developed so far, and are currently in use. Mainly, the forecasting techniques in an SC setting have been discussed from the perspective of BDA. This research aims to continually enhance the performance of a forecasting model incorporating a sustainable circular BDA-SCM framework that can drive future research using business intelligence and value theory as theoretical approaches. Systematic literature review (SLR) was used to perform this research, which is an approach used to locate, evaluate, and interpret what relevant research has been done on a specific issue or topic by sketching out and analyzing the current intellectual landscape \cite{tranfield2003}. The review paper attempts to combine the applications of BDA and ML forecasting in SCM by seeking the solutions to the following research questions (RQs) for guiding the study's development to accomplish its overall objective:
\begin{itemize}
    \item RQ1: What are the efficient steps to formulate an ML Forecasting model to predict the SC factors?
    \item RQ2: How can the forecasting, SC decision-making, and performance measurement processes be connected, tracked, and optimized in cyclic order?
    \item RQ3: How can forecasting affect SC performance, and which ML forecasting models are relevant to SC forecasting?
\end{itemize}

The scope of this research extends beyond the technical intricacies of forecasting algorithms. It delves into the broader implications of forecasting on the human workforce, inventory management, and the overall performance of the SC. By addressing the adverse effects of phantom inventory and emphasizing the dependency of managerial decisions on SC key performance indicators (KPIs), this research contributes to the overarching goal of improving operations management, transparency, and planning efficiency. The novelty of this paper lies in several key aspects:

\begin{itemize}
    \item The introduction of a comprehensive BDA-SCM framework that provides a holistic view of SC forecasting and highlights interconnections between processes, offering a novel perspective.
    
    \item The integration of ML techniques within SCM for forecasting purposes presents novel approaches to enhance accuracy and effectiveness.
    
    \item Addressing the issue of phantom inventory, providing insights and potential solutions to improve inventory management practices and forecasting precision.
    
    \item Exploring the connection between accurate forecasting and SC performance, offering a novel perspective on leveraging forecasting models for optimization.

    \item Conducting a comprehensive survey of 152 papers spanning several decades, providing a unique and valuable contribution to the field and consolidating a vast body of literature.

    Including papers from such a broad timeframe allows for identifying trends, shifts in methodologies, and key milestones in the field. This comprehensive survey sets this paper apart from other SLR review papers that have not undertaken such an extensive examination of the literature. The insights gained from this extensive survey enhance the robustness and reliability of the conclusions drawn in the paper.
    
\end{itemize}
The forthcoming sections will comprehensively explore the ``\hyperref[sec1]{Research procedures}," and the proposed cyclic connection embedded within the framework will be elucidated in ``\hyperref[sec2]{BDA-SCM framework}". Moving forward, ``\hyperref[sec3]{Pre-process}" section addresses critical components, including the imperative need for strategic data collection aligned with SC objectives, methodologies for data preprocessing, feature engineering (FE), exploratory data analysis, and the classification of forecasting types based on distinct time horizons. The ``\hyperref[sec4]{Control-process}" section delves into optimizing preprocessing methodologies. This optimization, rooted in post-process KPIs, aims to elevate overall control processes, encompassing inventory management, workforce determination, cost optimization, and production and capacity planning. The discussion on SC KPIs and error-measurement systems for model optimization unfolds in the ``\hyperref[sec5]{Post-process}" section. This section aims to provide insights into refining forecasting models for superior performance. In the ``\hyperref[sec6]{Challenges}" section, attention is directed toward acknowledging and addressing technological obstacles encountered during the extensive review of pertinent articles. This section highlights and discusses the challenges inherent in navigating the technological landscape within the scope of this research. In the ``\hyperref[sec7]{Practical Implications}" section, we delve into actionable insights for SC practitioners, detailing the implementation of the proposed BDA-SCM framework in real-world scenarios and outlining the substantial benefits they can expect. By incorporating these novel aspects, the paper contributes to the existing body of knowledge in the field of BDA-SCM framework for forecasting, offering new insights, methodologies, and recommendations for future research and practical implementation.

\section{Research procedures}
\label{sec1}
\subsection{Planning the review}
In this article, the BDA-SCM cyclic framework was initially developed to incorporate pre-process, control process, and post-process phases. Each phase was illustrated utilizing the most relevant selected works of literature. For forecasting purposes, pre-process recommendations include a step-by-step approach to forecasting and BDA best practices to facilitate comprehensive demand forecasting considering state-of-the-art technologies and relevant research. In the control process, how SC factors and forecasting affect workforce efficiency have been discussed. The post-process portion of the managerial decision-making process explains how managers use the KPI and optimization of the forecasting model to choose the appropriate metrics and insights.
\subsection{Conducting the review}
\subsubsection{Search strategy}
\begin{figure*}[!ht]
  \centering
  \includegraphics[width=0.7\textwidth, height=0.7\textwidth]{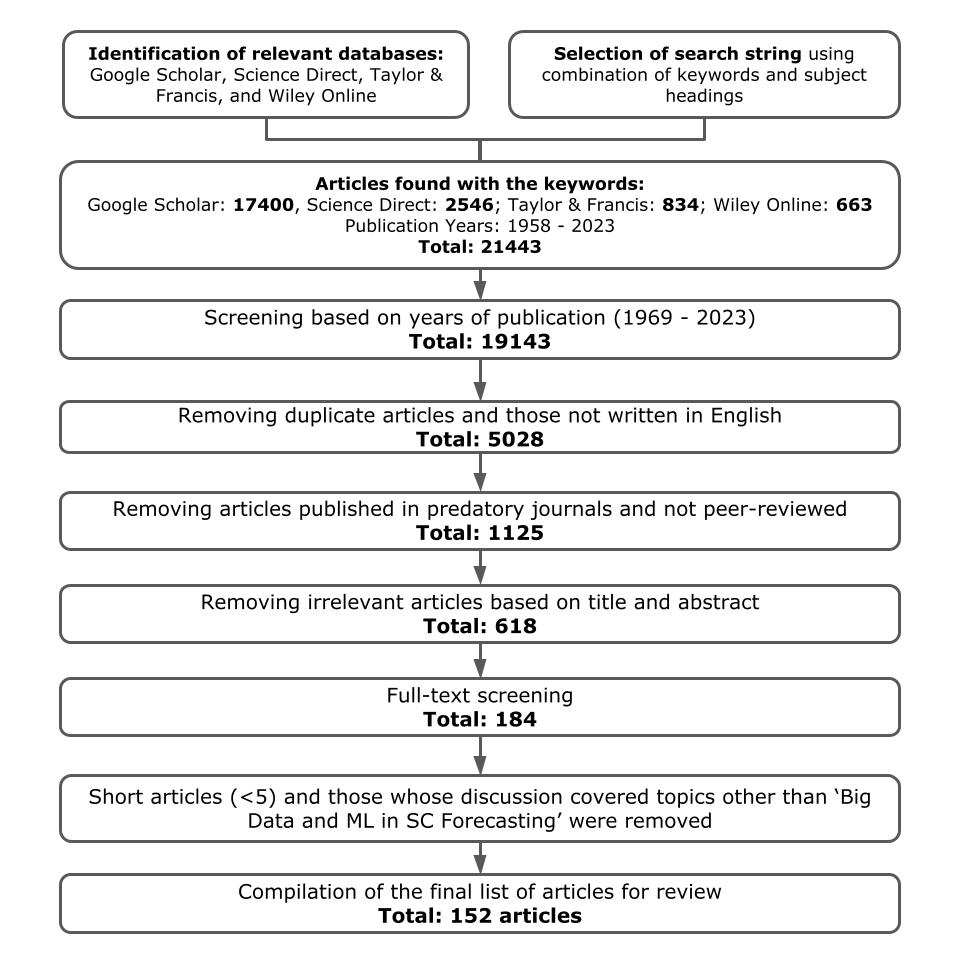}
  \caption{PRISMA flow diagram illustrating the article selection process for the SLR on BDA-SCM forecasting.}
  \label{fig1}
\end{figure*}

This SLR aimed to provide a comprehensive and objective evaluation of the existing research until 2023 on BDA-SCM, including an investigation and analysis of various SC forecasting problems and BDA innovations, strategies, and techniques. Major academic databases, including Google Scholar and Science Direct, were searched to minimize bias and ensure the inclusion of a broad range of relevant sources and content. Only English articles published in peer-reviewed journals in the fields of Computer Science, Business, Management and Accounting, Engineering, and Decision Sciences were included. Figure \ref{fig1} shows the PRISMA flow diagram for the systematic review process, which includes the number of articles identified, screened, and included in the analysis.
A combination of keywords and subject headings related to the topic of interest was used to develop the search strategy. The search strings were limited to the title, abstract, and keywords fields and included the following terms:
\begin{itemize}
    \item (“Data Analytics” OR “Big Data” OR “Data Analysis”) AND (“Supply Chain Management”) AND (“Forecasting”) 
    \item (“Data Preprocessing” OR “Data Wrangling” OR “Supply Chain Data Analysis”)
    \item (“Supply Chain Forecasting” OR “Demand Forecasting”)
    \item (“Warehouse” OR “Inventory”) AND (“Workforce” OR “Human”) AND (“Forecasting”)
    \item (“Supply Chain Performance” OR “Supply Chain KPI” OR “Supply Chain Monitoring”)
    \item (“Forecasting KPI” OR “Forecasting Error Measurement” OR “Forecasting Performance”)
    \item (“Forecasting Model” OR “Time-series Forecasting”)
\end{itemize}

\subsubsection{Selection strategy}
The relevance of each publication was assessed to ensure that the selected papers were empirically sound and conceptually relevant to BDA-SCM-related research advances. Articles were considered more relevant if the search terms appeared in the title, abstract, keywords, and throughout the text. The identified papers were critically analyzed, particularly regarding the relevant sections that mentioned BDA-SCM. This approach drew from relevant views on SCM-forecasting challenges and BDA techniques and helped to achieve the research review goals.
The remaining articles were then assessed to verify that they provided the necessary research perspective and empirical data to meet the review's objectives. Finally, to ensure that the selected articles aligned with the review goals, we conducted a rigorous alignment process, comparing the articles to the research review objectives. Only articles that met all of the selection criteria were included in the final review.

\section{BDA-SCM framework}
\label{sec2}
\begin{figure*}[!ht]
  \centering
  \includegraphics[width=1\textwidth]{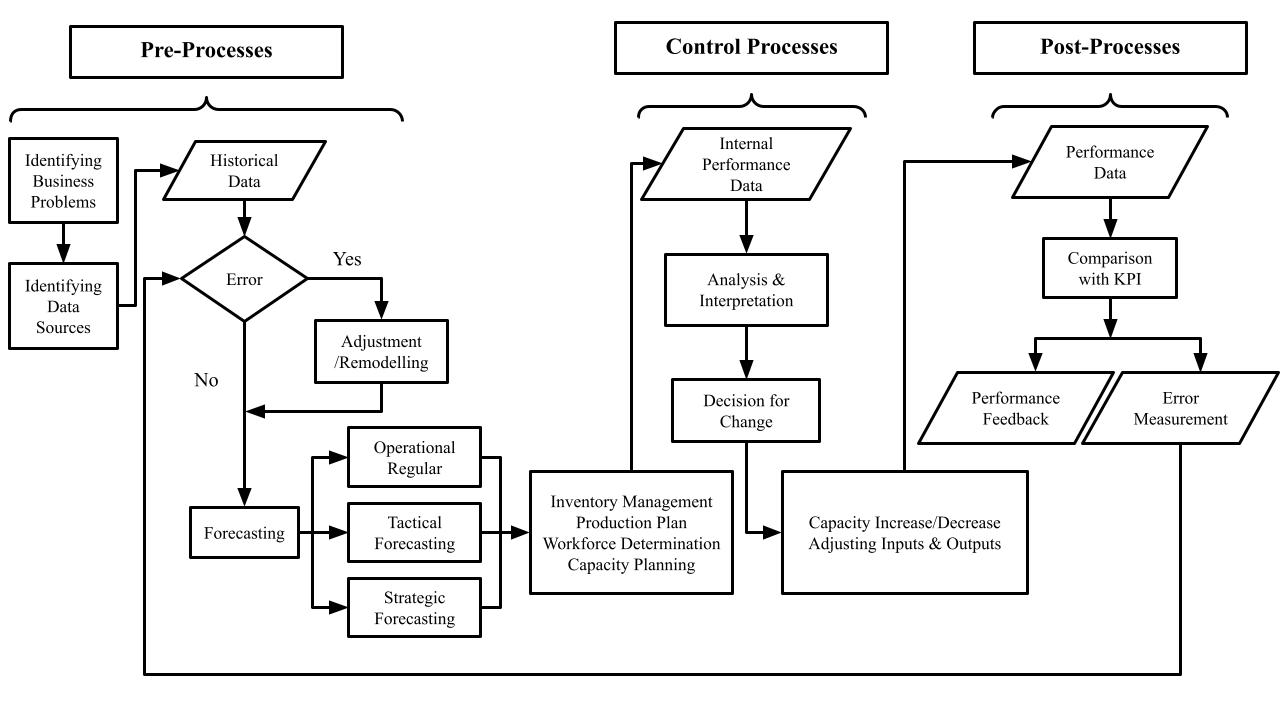}
  \caption{Big data analytics in supply chain processes (pre-process, control-process, post-process).}
  \label{fig2}
\end{figure*}

\begin{figure*}[!ht]
  \centering
  \includegraphics[width=1\textwidth]{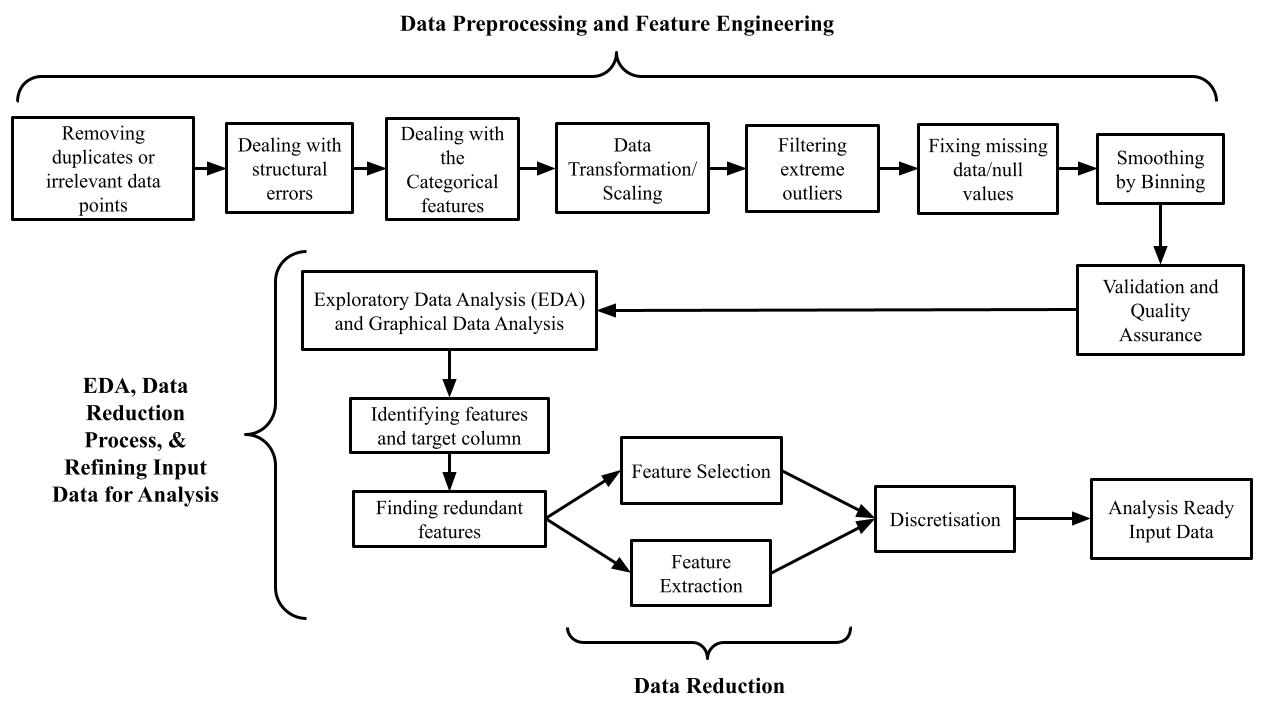}
  \caption{Data preprocessing, feature engineering, exploratory data analysis, and data reduction.}
  \label{fig3}
\end{figure*}

\begin{figure*}[!ht]
  \centering
  \includegraphics[width=1\textwidth]{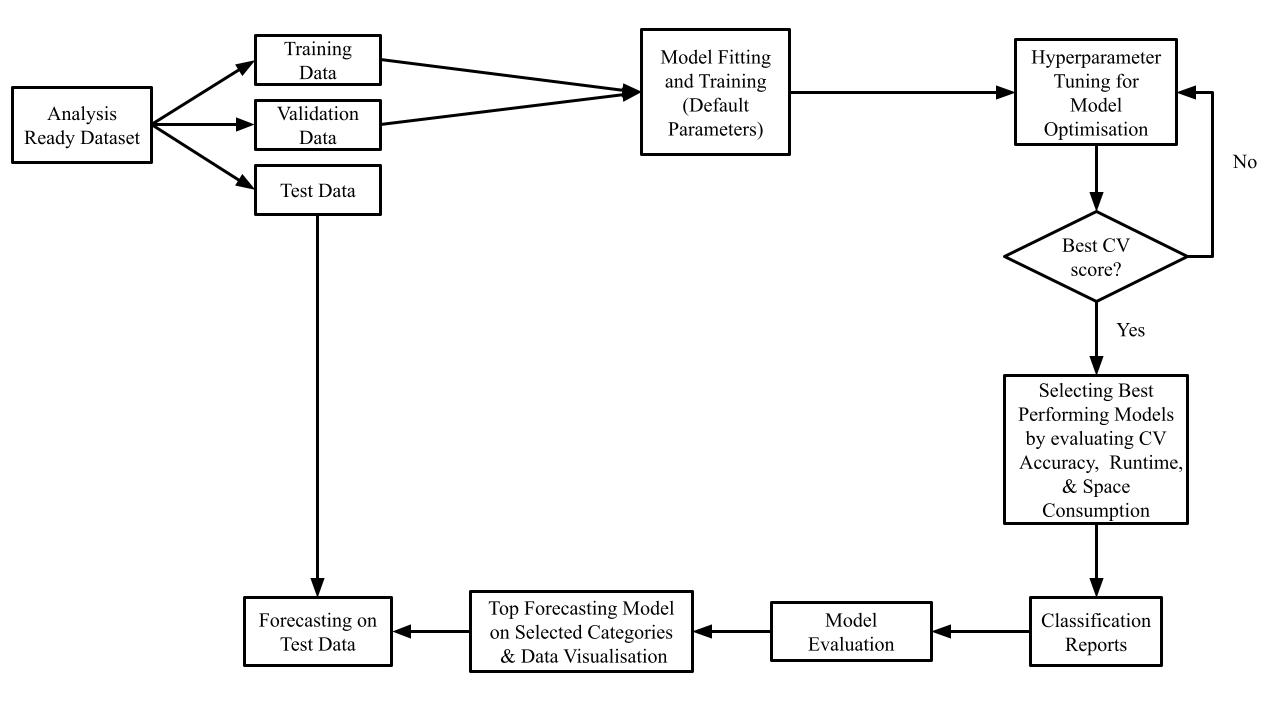}
  \caption{Machine learning model training, hyperparameter optimization, and model evaluation.}
  \label{fig4}
\end{figure*}

The proposed BDA-SCM framework, depicted in Figures \ref{fig2}, \ref{fig3}, and \ref{fig4}, establishes a cyclic connection that facilitates continuous improvement in SC forecasting. This cyclic process seamlessly integrates three essential stages: Pre-process, Control-process, and Post-process, fostering a dynamic relationship that optimizes SC operations iteratively. Figure \ref{fig2} mainly consists of the use and cyclic flow of data in SC. It only includes the SC parts where BDA may be involved. Figure \ref{fig3} complements Figure \ref{fig2} by mentioning the methods for cleaning, exploring, and analyzing data properly. It includes FE techniques to select only the most relevant and unique features from which ML algorithms can learn efficiently. Finally, Figure \ref{fig4} is a proposed method for data splitting, model training, hyperparameter optimization, cross-validation, testing, and evaluating errors to perfect the forecasting methods mentioned in Figure \ref{fig2}.

In the Pre-process stage, the focus is on ensuring accurate and relevant data aligned with SC objectives. The cyclic nature of this stage involves a continuous feedback loop. For example, after training an initial ML forecasting model, the performance is evaluated using real-time data. Any discrepancies or deviations from expected outcomes trigger a revisit to the Pre-process stage. This might involve reassessing data collection methods, exploring new data sources, or refining the preprocessing steps to enhance the quality of input data. The Control-process stage benefits from the cyclic connection, encompassing decision-making areas like production planning, workforce determination, and inventory management. Suppose a decision made based on forecasted data results in suboptimal outcomes. In that case, this feedback loops back to the Pre-process stage. The system may reevaluate the forecasting model's inputs, incorporating real-time data to enhance decision-making accuracy in subsequent cycles. In the Post-process stage, the cyclic connection enables continuous performance improvement. After the initial model predictions, performance metrics are analyzed, and any deviations from expected results trigger a reevaluation of the forecasting model. This feedback loop, integrated into the Post-process stage, ensures that the model evolves over time, adapting to changing SC dynamics and improving its predictive capabilities.

Consider a scenario in demand forecasting where the initial ML model predicts a surge in demand for a particular product. If the actual demand deviates from the forecast, the cyclic connection triggers a reassessment in the Pre-process stage. Analysts may explore new data sources, refine data preprocessing methods, or adjust FE techniques to capture changing demand patterns more accurately. In inventory management, the Control-process stage involves decisions on stock levels based on forecasted demand. If the actual inventory levels deviate significantly from the forecast, the cyclic connection prompts a revisit to the Pre-process stage. This may involve refining the preprocessing of inventory data, incorporating real-time data on SC disruptions, or adjusting the forecasting model to enhance inventory optimization. In production planning, the decision-making process relies on accurate product demand forecasts. If the actual production output falls short or exceeds the forecasted demand, the cyclic connection triggers a reassessment in the Pre-process stage. This may involve refining data collection methods, exploring new features relevant to production efficiency, or adjusting the forecasting model to better align with dynamic production needs.

\section{Pre-process}
\label{sec3}
\subsection{Identifying business problems}
At the outset, the type of data that needs to be collected, stored, analyzed, and interpreted is selected based on SC strategies. \cite{wieland2013} categorized SC strategies based on risk and impact, such as robustness for low-impact high-risk, agility for low-risk high-impact, rigidity for low-impact low-risk, and resilience for high-impact high-risk decisions. Varying levels of responsiveness and efficiency can adopt the strategies. Responsiveness has been a critical factor in gaining a competitive advantage, and it depends on the deviations in demands and a company's capability to respond to such deviations. An increase in responsiveness decreases efficiency, and vice versa \cite{minnich2006}. The responsiveness level affects product volume, order fulfillment rate, workforce, manufacturing capacity, warehouse capacity, transportation carriers, product mix, supplier’s product mix, inbound and outbound logistics, etc. \cite{thatte2013}. Therefore, whether data needs to be collected should be decided based on responsiveness, as different data sets are required to boost the SC efficiency and responsiveness by allocating them. Furthermore, the frequency of data analyses would also depend on the responsiveness level of the company. In short, the factors that may dictate the sort of forecasting required for a business include the context of forecasting, the types of data available, the required level of accuracy, the length of the forecasting period, the time available for each forecast, and the value addition made through the forecast \cite{chambers1971}.

\subsection{Identifying data sources}
Once the data that needs to be gathered is selected, identifying the sources is essential. Determining variables is required for timely forecasts to bring helpful information \cite{duarte2006}. Moreover, a conclusion may not be based on a single type of data; the initial conclusion can be validated based on multiple data types. \cite{varela2014} mentioned 56 different data sources for four main SCM levers, procurement, warehouse operations, marketing, and transportation, as leveraging various data sources allows finding actionable insights quickly; some of the more relevant data sources have been listed below:

\begin{enumerate}
  \item Transportation
  \item Barcode systems
  \item Demand chain
  \item CRM Transaction data
  \item BOMs
  \item Customer surveys
  \item Blogs and news
  \item Demand Forecasts
  \item Procurement
  \item Delivery times and terms
  \item Invoice data
  \item ERP Transaction data
  \item GPS-enabled BD telematics
  \item Product reviews
  \item Competitor pricing
  \item Inventory costs
  \item Customer Location and Channel
  \item Traffic density
  \item Email records
  \item Crowd-based Pickup and Delivery
  \item Equipment or asset data
  \item Intelligent Transport Systems
  \item EDI purchase orders
  \item Warehouse operations
  \item Logistics Network Topology
  \item In-transit Inventory
  \item SRM Transaction data
  \item Transportation Costs
  \item Warehouse Costs
  \item Pricing and margin data
  \item RFID
  \item Origination and destination (OND)
  \item Local and global events
  \item Supplier current capacity \& customers
  \item Sales history
  \item Weather data
  \item SKU level
  \item Supplier financial performance information
  \item Raw material pricing volatility
  \item On-Shelf-Availability
  \item P2P (Procure-to-Pay)
  \item Product traceability \& monitoring system
\end{enumerate}

\subsection{Data preprocessing and feature engineering (FE)}
\subsubsection{Duplicates removal}
It is problematic to waste space and runtime with duplicate rows. Duplicate rows create incoherence, and the ML model fails to learn new information. Because of input mistakes, changes in some feature values (e.g., the identifier value) may generate duplicate rows that will be deemed distinct by the machine. It is easier to drop the duplicates or substitute them with relevant values using data preprocessing libraries in Python and R languages. Nevertheless, the main challenge is identifying factors on which the duplicates should be removed. Of the number of methods invented to remove duplicates, we review the following:

\textit{Bayesian:} The Fellegi-Sunter-algorithm is the most commonly used model in probabilistic approaches because of its Bayesian nature \cite{fellegi1969}. The Bayes Decision Rule is a common approach \cite{elmagarmid2007}. A Bayesian inference difficulty may develop when the probability density of a unique row differs from a duplicate record, and the functions are known. Neural network (NN) algorithms are more accurate without the Fellegi-Sunter-algorithm if the data are adequately described or labeled \cite{wilson2011}.

\textit{Partitioning Methods:} Clustering methods identify and drop duplicates utilizing graph partitioning approaches \cite{singla2004}. However, \cite{hassanzadeh2009} compared 12 clustering methods and found that the popular sophisticated algorithms provided lower accuracy, first suggesting that Markov Clustering is a more scalable, accurate, and efficient algorithm.

\textit{Aggregate fitting:} CART \cite{cochinwala2001} and SVM \cite{joachims1999} aggregate fitting results for various row features. SVM is highly memory efficient and works well with lots of dimensions. However, it does not work well with large datasets or data with overlapping classes. CART is intuitive and easily used. The problem is that data is classified based on the sample and may not apply to larger datasets.

\textit{Others:} Bootstrapping clusters \cite{verykios2000} or hierarchical graph structures encode the features as non-matchable binary features creating dual probability densities rather than probabilistic distribution modeling for the inspected quantities \cite{ravikumar2012}. Bootstrapping clusters are used for unsupervised data. Simple techniques have long been studied, such as utilizing distance measurements to identify duplication \cite{monge1996}. Weighted transformations also occur in literature \cite{dey1998}. Additional methods, like ranking the most same-type weighted rows comparable to those provided, are also utilized to identify the least duplicated rows \cite{guha2004}.

\subsubsection{Dealing with categorical features}
One-hot, ordinal, Helmert, polynomial, and binary encoders are outperformed with a 95\% accuracy by Sum, and Backward Difference encoders are preferred for prediction jobs \cite{potdar2017}. \cite{lopez2020} presented a generic Information-based encoder that transforms mixed-type features into numeric ones while maintaining the dataset’s original dimension, with better accuracy than One-Hot and Feature-Hashing. \cite{garnier2019} demonstrated that Ordinal-encoder (straightforward and convenient to execute but incorporates a sequence of features) outperformed Hashing (introduces limited features and moderately ignores the feature sequence); One-hot-encoding generates a massive number of features and forces the use of a very simplified regression analysis. To train residual features from time categorical variables derived from variable time stamps, a DeepGB neural network with embedding layers may be used, which are necessary to learn multiple time series at once to encode categorical features in a lower dimension or by embedding their IDs and retrieve helpful information \cite{karingula2021}.

\subsubsection{Data scaling}
Normalization is valuable when using ANN, clustering techniques, or classification software. The learning phase may be accelerated by normalizing the data features in tanning faces for backpropagation NN methods.

\textit{Min-max normalization:} The scaling of $b$ values of a numerical feature $F$ to a defined range represented by $[\text{new-min}_F, \text{new-max}_F]$ is termed min-max normalization. To acquire the new value, the following equation is applied in $b$ to produce a changed value $b'$:

\begin{equation}
\small
b' = \frac{{(b - min_F)}}{{max_F - min_F}} \cdot (new - max_F - new - min_F) + new - min_F
\label{eq1}
\end{equation} where $max_F$ and $min_F$ mean the maximum and minimum feature values, respectively. In normalization, $[new-min_F, new-max_F] = [0,1]$ or $[-1,1]$ are the usual intervals \cite{bib5}.

Datasets prepared for use with distance-based learning methods commonly use this normalization technique. The features having a significant $max_F-min_F$ difference will be prevented from dominating the distance computation by applying a normalization to rescale the data to the same value ranges, and it will not be able to distort the learning process by assigning the older features much weight. It is also known to help ANNs learn faster by allowing the weights to converge more quickly. 

\textit{Z-score normalization:} Min-max normalization is not practicable if the minimum and maximum values are not provided. Even when these values are known, the existence of outliers might cause the min-max normalization to be skewed by clustering the values and restricting the computational accuracy available to represent them.

\begin{equation}
b' =  \frac{(b - \bar{x})}{s_x}
\end{equation}
where $\bar{x}$ is the sample mean.
\begin{equation}
\bar{x} = \frac{1}{n} \sum_{i=1}^{n} b_i
\end{equation}
Moreover, $s_x$ is the mean absolute deviation of $x$ \cite{han2012}.
\begin{equation}
s_x  =  \frac{1}{n} \sum_{i=1}^{n}(b_i-\bar{x})
\end{equation}

\textit{Decimal scaling normalization:} Normalising the numerical feature values by relocating the decimal-point by 10th power divisions so that the $\text{$highest$ $absolute$ $value$}<1$ after transformation is a simple method for minimizing the absolute feature values.
\begin{equation}
b' =  \frac{b}{10^k}
\end{equation}
where $k$ is an integer (the lowest), such that $new-max_F<1$.

\subsubsection{Data transformation}
Data transformation can create new features, also known as changing features, where mathematical formulae derived from business models or pure mathematical formulae are used to integrate the raw input features. Linear, quadratic, polynomial, non-polynomial, rank, and Box-Cox transformations are a few of the different existing transformation techniques. 

Normalizations may not be sufficient in research experiments, and full automation to fit the data and optimize the resulting model. Combining the data embedded in several features may be advantageous in some circumstances. Linear transformation based on simple algebraic operations is a basic approach that may be utilized for this goal. A quadratic transformation can occur when a newly introduced feature is formed using the expressions in quadratic form. Using the fundamental features of the dataset, quadratic modifications can assist us in uncovering information that is not directly there. Transformation approximation using polynomials could be implemented by brute force exploration with one unit at a time when no expert assistance can tell us which transformation and features to employ. The transformation of the rank approach is recommended for identical training and test data or a complete dataset for DA and cluster analysis model development \cite{refaat2010}.

Nonparametric approaches using rank transformation are not recommended to be introduced into traditional statistics courses because it inhibits how widely the nonparametric technique may be used, which is unnecessary. Another misperception is that the nonparametric technique is utilized chiefly for hypothesis testing. This entirely obscures the superior theoretical and conceptual flexibility of many nonparametric methods.

\cite{spitzer1978} studied the limited sample aspects of the estimated parameters using the Box-Cox transformation. Under the premise of approximating normalcy, the technique worked well. The outputs were utterly impartial for forecasting, and their differences were surprisingly small. Asymptotic variances and stability features of Box-Cox estimates in the linear model were examined by \cite{bickel1981}. In the case of unknown transformation parameters, linear regression models with minor to intermediate error variances showed much higher asymptotic variances than known ones. Furthermore, they observed that Box-Cox approaches perform inconsistently in models with minor to intermediate residual variance.

\subsubsection{Filtering extreme outliers}
The most often recommended approach in the literature of outlier identification and repair is via filtering.

\textit{Outlier detection:} Statistical methods for detecting outliers include box plots, scatter plots, z-scores, and IQR (Interquartile Range) scores. Normal distribution empirical relations should be followed for outliers where the values are $<\mu$ - $3\sigma$ or $>\mu+3\sigma$ for normal distribution, where $\sigma$ and $\mu$ are the standard deviation and mean of a particular feature. IQR proximity rule should be used in which outliers are $<(Q1$ – $1.5\times IQR)$ or $>(Q1 + 1.5\times IQR)$ for skewed distribution. For other distributions, a percentile-based approach should be used in which values that are distant from the 99 percentile and $<1$ percentile are regarded as outliers.

\textit{Outlier Treatment:} Various techniques can be employed to address outliers within a dataset. Trimming, the first method, involves the removal of outliers, but it is generally not recommended due to potential information loss. As the second approach, capping identifies outliers based on a predefined threshold, either greater or less than the established limit. The number of outliers in the dataset influences the determination of this capping threshold. Alternatively, outliers may be treated similarly to missing values (MVs). Lastly, Outlier Removal Clustering (ORC), a modification of K-Means Clustering, eliminates outliers in iterative loops. ORC effectively removes outliers from clusters, and careful parameter adjustments are essential as the dataset influences model precision. Importantly, ORC ensures that the computation of centroids remains unbiased, particularly when dealing with distant locations from the k-clusters.

\subsubsection{Dealing with missing values (MVs)}
In SC Data Analysis, one of the preprocessing techniques, Imputation, is adopted to overcome the drawbacks of MVs. The most straightforward approach to drop the rows having MVs is if a comparatively small fraction of observations is present and the analysis of all rows is not substantially skewed in interpretation \cite{little1987}. \cite{barnard1999} showed that MVs are generally connected with three sorts of issues:
\begin{enumerate}
    \item Inefficiency
    \item Difficulties in managing and interpreting data
    \item Skewness because of discrepancies between perfect and missing data.
\end{enumerate}

When it comes to MV therapy, there are generally three options \cite{farhangfar2007}:
\begin{enumerate}
    \item First, eliminate all instances that have MVs in their features. Thus, removing features with higher-than-normal MV levels falls within this area.
    \item When estimating the model parameters for a whole dataset, another way is to employ maximum likelihood processes, using the obtained model parameters for imputation via sampling.
    \item Finally, MV imputation is a group of processes focused on substituting predicted MVs for existing ones. Most of the time, the features in a data set are interdependent. As a result, MVs may be calculated by identifying correlations among features.
\end{enumerate}

\textit{Common approaches:} To keep the MVs unchanged, known as Do Not Impute (DNI), is the most straightforward approach where if the baseline MVs strategies are available, the algorithm must employ them. When many rows include MVs and using DNI would lead to an irrelevant, inaccurate, and small dataset, then MVs are commonly substituted by the universal-most-frequent feature value for nominal features and the universal mean value for quantitative features \cite{grzymala2005}. \cite{batista2003} showed another process utilizing Hot Deck that partitions the complete dataset into clusters, links each row with a cluster, and fills up the MVs, where any complete row from the cluster can be utilized. The imputation of Cold Deck is identical to the hot deck, except the dataset cannot be the existing dataset. They demonstrated that the MVs imputation based on the KNN might beat the internal techniques assessing C4.5 and CN2 to handle MVs and exceed the imputation method of mean or mode, which is widely intended to treat MVs.

\textit{Maximum likelihood imputation methods:} Assume for $n$ independent rows $(i=1,\ldots,n)$, there are $k$ variables $(y_i1,y_i2,\ldots,y_ik)$ with no missing data. The maximum likelihood function is:
\begin{equation}
\small
    L = \prod_{i=1}^n f_i(y_{i1}, y_{i2}, \ldots, y_{ik}; \theta)
\label{eq6}
\end{equation}

Assume that $y_1$ and $y_2$ have MVs that fulfill the Missing at Random (MAR)-assumption for a specific row $i$. The combined probability for that observation is the chance of witnessing the remaining features, $y_{i3}$ $ $ through $y_{ik}$. If $y_1$ and $y_2$ are two discrete features, this is the aforementioned combined probability multiplied by all potential values of the two features with MVs:
\begin{equation}
\small
f_i^* (y_{i3},\ldots,y_{ik};\theta) = \prod_{y_1}\prod_{y_2}f_i(y_{i1},\ldots,y_{ik};\theta)
\label{eq7}
\end{equation}

For continuous MVs,
\begin{equation}
\small
f_i^*(y_{i3}, \ldots, y_{ik}; \theta) = \int_{y_1} \int_{y_2} (y_{i1}, y_{i2}, \ldots, y_{ik}) \, dy_2 \, dy_1
\label{eq8}
\end{equation}

The multiplication of probabilities for all the rows is the overall likelihood. If there are $q$ rows with full data and $p-q$ rows with MVs on $y_1$ and $y_2$ features, then the ML function becomes:
\begin{equation}
\small
L = \prod_{i=1}^q f_i(y_{i1}, y_{i2}, \ldots, y_{ik}; \theta) \prod_{i=q+1}^p f_i^*(y_{i3}, \ldots, y_{ik}; \theta)
\label{eq9}
\end{equation}

\cite{bib2} narrowed down the following imputation options using non-parametric statistical testing:
\begin{itemize}
    \item Row elimination (IM) and no imputation (DNI) methods are outperformed by imputation techniques that fill in the MVs.
    \item No single-size/generic imputation method works for all regressors or classifiers.
\end{itemize}

CMC and EC methods are proposed that yield a lower noise ratio for Wilson and balance the mean MI difference. The proposed imputation approaches focused on classification techniques, including Rule Induction Learning Models: FKMI, Black Boxes Methods: EC, and Lazy Learning (LL) models: MC.

\subsubsection{Binning}
In this method, a continuous variable is converted into a group of intervals. Each interval can then be treated as a ‘bin,’ with the option of enforcing an order dependent on the data's subsequent processing. While smoothing, each bin's min and max values are calculated as bin borders. Then, for each value, the nearest border value is substituted. Typically, the smoothing effect increases with bin width. If the bin widths are identical, binning may be used as a discretization method by substituting mean or median for bin value. It is possible to create hierarchical ideas by iterating over this procedure indefinitely. It's unsupervised since class labels are not used, and the user specifies bin numbers.

\subsubsection{Deep learning (DL) based FE}
A multi-filter NN (MFNN) end-to-end model was developed for multivariate financial time-series FE and classification-based forecasting utilizing DL techniques \cite{peng2005}. Their proposal MFNN was 15.41\% higher than the best result (Logistic Regression) of traditional ML models and 22.41\% higher than the statistical approach (Linear Regression) in terms of returns.

\subsection{Exploratory data analysis (EDA) \& data reduction (DR)}
In this process, the target or dependent column and independent features are obtained. The DR, EDA, and clustering techniques reduce runtime and space during the deep-learning modeling phase. DR can be employed to decrease the size of a dataset while still keeping the data’s original integrity. In our framework, we suggest performing Feature Selection and Feature Extraction simultaneously after selecting the target column and finding redundant features; then, Discretisation may be performed if necessary. Then, the dataset will be ready for further analysis and model training.

\subsubsection{Identifying redundant features}
Feature Redundancy lengthens the modeling time of ML algorithms and leads to model overfitting. Feature redundancy arises from the possibility of derivation from another feature or set of features. The following techniques may be adopted to handle redundancy:

\textit{Covariance and correlation:} In statistics, covariance refers to the amount that two features or factors change in tandem whose value lies in the $(-\infty, +\infty)$ range. Positive covariance indicates they move in the same direction. Negative covariance means that any features are greater than the mean, and others are less than the mean, and vice-versa. Zero covariance means features may be independent under a certain hypothesis. 
On the other hand, correlation analysis is a widely used dimensionless measurement ranging from $-1$ to $+1$ to discover redundancies in numerical features that evaluate and quantify the relationship intensity. The features are positively correlated for correlation values greater than zero (0); for zero, they are independent; for less than zero, they are negatively correlated \cite{bib5}. Covariance and correlation are directly proportional to each other. 
In numeric feature selection, correlation is better to use, as correlation analysis is scaled [-1, 1], but the covariance range is indefinite $(-\infty, +\infty)$. We should choose correlation for better interpretation. Changes in location, size, or scale have no effect on correlation. However, both of them are limited to only being able to identify linearity.

\textit{$\chi^2$ correlation:} The $\chi^2$ (Chi-Square) test is often used when dealing with nominal features and finite value sets. We can use the $\chi^2$ test to see whether there is any link between the values of two nominal features, where a probability table with joint events is established. If $\alpha$ (significance level) is less than the estimated one (or the $\chi^2$-value (calculated) > table value), the null hypothesis gets discarded, and the two features can be said to be correlated statistically \cite{bib5}.
SC Analysts must remember that the $\chi^2$-test does not tell much about the strength of the relationship between two features. The $\chi^2$-test offers advantages such as resilience regarding data distribution, computational simplicity, extensive information produced from the test, utilization for investigations where parametric criteria cannot be satisfied, and scalability in processing data from two and multiple-group research. The drawbacks are sample size constraints and difficulty in comprehension when there are many (>20) features.

\subsubsection{Feature selection (FS)}
The reasons for conducting FS may include removing unnecessary data, enhancing forecasting accuracy, reducing data cost and model complexity, and improving training efficiencies such as reductions in space needs and computational costs \cite{saeys2007}. FS approaches, despite their widespread use, have several drawbacks:

\begin{itemize}
    \item Training data size significantly impacts the subsets produced by many FS models (particularly those created using wrapper-based techniques). If the training data is limited, then the feature subsets retrieved will be limited, resulting in the loss of key variables.
    \item Because the target feature is connected with many independent features, and their removal would adversely influence learning accuracy, reducing high-dimensional data to a limited range of features is not always possible.
    \item When dealing with huge datasets, a reverse elimination approach takes too long since the algorithm must make judgments based on enormous amounts of data in the early stages.
    \item In certain circumstances, FS results will still include significant important features that may obstruct the use of complicated training strategies.
\end{itemize}

\textit{Leading methods:} In order to create FS techniques by combining a feature evaluation score and a cutting criterion, \cite{arauzoazofra2011} recommended that functions based on information principle produce better accuracy, not suggesting any universal cutting condition. However, those independent of the metric perform best, and outcomes differ across models. For each kind of model, wrapper techniques were recommended to avoid this effect. 

\cite{bolon2013} investigated nine feature selectors running across 11 simulated datasets to examine the methodologies in the context of a growing number of unnecessary features, noise in the data, redundancy, correlation between attributes, and the ratio of observations to features. ReliefF proved to be the best alternative regardless of the specifics of the data, and it is a filter with a cheap computational expense. Wrapper techniques have proven to be an intriguing choice in specific disciplines if they can be used with the same classifiers and consider the greater computing costs. Extensive theoretical research has been conducted on the Relief and its variants, showing that they are resilient, noise-resistant, and can decrease their space-time complexity in parallel \cite{robnik2003}. 

Since the emergence of rough sets in pattern recognition, several FS techniques have based their criteria for assessing reductions and approximations based on this idea \cite{swiniarski2003}. Because total searches of substantial datasets are impossible, stochastic methods based on meta-heuristics and approximate assessment criteria have also been explored. \cite{wang2007} utilized particle swarm optimization for this job. Features are discontinuous, making it challenging to pick approximately set-based characteristics in the literature. Rough set-based feature selection's key drawback is the constraining condition that all values be discrete, for which issue, a fuzzy rough FS method (FRFS) was suggested \cite{jensen2007}.

When data is vast, messy, blended with categorical and numerical variables, and may have dynamic effects requiring sophisticated models, the synthesis of forecasting analytics in the form of ensembles can create a compressed sample of non-redundant features \cite{tuv2009}. There are four phases to the technique suggested here: identifying relevant features, computing masking scores, removing the masked factors, and generating residuals for progressive modification. The Random Forest ensemble is considered in all four stages.

Two problems arose simultaneously with the growth of highly-dimensional data: FS is essential in every training, and the accuracy and robustness of the FS algorithms may be ignored. \cite{rodriguez2010} discussed the FS reduction job introducing the Quadratic Programming FS (QPFS), which utilizes the Nyströn-approximation-matrix diagonalization method for large datasets. mRMR and ReliefF were outperformed using Pearson's correlation coefficient and MI. A local learning-based approach may be beneficial when assessing many irrelevant attributes and complicated data ranges \cite{sun2010}. The impacts of high-dimensional datasets may be mitigated by pre-processing the feature ranking procedure to exclude class-dependent density-based features \cite{javed2012}. To scale any method in significant data issues demands cutting-edge distributed-computing frameworks like MapReduce and Message Passing Interface (MPI) \cite{zhao2013}.

We can use supervised FS if the data has class labels; otherwise, unsupervised FS is the best option. This approach generally maximizes clustering efficiency or the FS based on correlation, feature dependency, and priority. The primary premise is to eliminate features that bring almost no value beyond what is already provided by the existing features in the system. \cite{mitra2002} suggested using feature dependency/similarity to reduce redundancy without needing a search procedure. An information compression metric called the maximum information compression index governs the clustering partitioning process, which uses features as the measure of similarity. Forward orthogonal search (FOS) is another unsupervised FS approach that aims to maximize the total reliance on the data to find relevant features \cite{wei2007}. Without compromising performance in clustering, the unsupervised FS used the Random Cluster Ensemble framework to compress the set of features by roughly 1/100 of its initial dimensions \cite{elghazel2015}. When compared to well-known classifications, precision/recall analyses revealed that feature weighting was highly successful in discovering the most suitable clusters \cite{modha2003}.

\subsubsection{Feature extraction}

Feature extraction accelerates the ML algorithm’s execution, optimizes raw data quality, boosts the algorithm’s efficiency, and simplifies the interpretation of the findings.

Principal component analysis (PCA):  It aims to analyze a collection of features’ variance-covariance patterns employing a few linear combinations and seeks the optimal $k$ number of $n-dimensional$ orthogonal vectors for data description, where $k \geq n$. Accounting for the most critical percentage of the discrepancy in the original dataset, the principal component (the first derived feature) is produced in decreasing order of contribution. Typically, for containing $\geq$ 95\% variance, just the top few principal components are retained. PCA is beneficial when many independent variables correlate with one another \cite{johnson2007}. The principal component is quick and comprehensive and ensures a solution is found for all datasets \cite{bib3}.

\textit{Factor analysis:} The fundamental concept underlying component analysis identifies a collection of influencing factors to restore the current features through a series of linear adjustments on the components. It is a method that finds out the range of factors along with their associated loadings, providing the features as well as the mean of the features \cite{johnson2007}. The factor models can be solved by (1) the Maximum-likelihood method and (2) the Principal-component method. Maximum likelihood presupposes actual data following a normal distribution and is computationally costly. 
The comparative differences between PCA and factor analysis are:
\begin{enumerate}
    \item Factor analysis, unlike PCA, implies a basic structure that connects the factors to the empirical observations.
    \item A three-factor system is substantially different from a two-factor system in factor analysis; however, in PCA, the two initial principal components stay the same when employing a third component.
    \item PCA is simple and quick. There are several methods for doing the computations in factor analyses, some of which are complex and tedious. 
    \item Using a sequence of linear transformations, PCA attempts to spin the original features’ axis. Again, factor analysis generates a new range of features to demonstrate the observed covariances and correlations.
\end{enumerate}
	
\textit{Multidimensional scaling (MDS):} MDS may be used in SCM to estimate the map depicting transportation distances between or within inventories using the distance matrix. The result is skewed owing to the disparity between calculated distances and the actual distances between inventories lying in a straight line. The map is typically centered on the origin and expanded to cover considerable distances. However, the answer may be found in any rotation.
\textit{Locally linear embedding (LLE):} With LLE, local linear fits are used to restore universal nonlinear configuration \cite{roweis2000}. All points are a linear weighted sum of their neighbors if adequate data is available. It is the basic notion behind the manifold approximation algorithm. For the LLE algorithm, the geometric principle is all that is required. LLE's advantages are that local minima are not involved in optimizations and have only two parameters. The embedded space has a universal coordinate system and preserves the local geometry of high-dimensional data. LLE also has several inherent shortcomings, which are stated as follows:
\begin{itemize}
    \item LLE generates folds and nonhomogeneous warps when the dataset is small or the points are irregularly measured.
    \item Noise significantly affects LLE, which causes embedding derivation errors.
    \item Short circuits may develop during the neighbors since the query typically uses Euclidean distance.
    \item Poor eigenproblems may arise.
    \item If two high-dimensional space observations differ, LLE cannot assure that their corresponding low-dimensional space instances also differ.
    \item LLE's embedding findings are extremely susceptible to its two system parameters: the number of clusters of each instance and regularisation.
    \item LLE presupposes that complete data exists on a unified surface and is unsupervised, but that does not happen for multi-labeled classification tasks.
    \item It is unclear how to assess the new sample data points because LLE does not provide a parameterized function that reflects between high-dimensional space and low-dimensional manifold. 
\end{itemize}

\subsubsection{Cardinality reduction}
The merging of two or more nominal or ordinal variables into a single unique category is called cardinality reduction. It is challenging to manage nominal features with a large number of groups. Converting high cardinal variables into binary variables provides many new variables, mostly zeroes. However, if utilized without conversion with models like Decision Tree that can accept them, there are risks of model over-fitting. So, decreasing the number of groups should be considered \cite{bib3}.

\subsubsection{Discretization}
The discretization method turns the numerical data into qualitative data, i.e., quantitative features, into discontinuous or nominal features that provide a non-overlapping segmentation of a linear system. Discretization can decrease data since it converts data to a much smaller sub-ensemble of discrete values from an enormous range of numerical values. Numerical features should be discretized as real-world dataset features are generally continuous, whereas most of the current ML algorithms can only be trained by utilizing nominal features in categorical data \cite{bib4}. 

Discretization generally involves four steps: (1) Continuous feature values to be discretized need sorting, (2) identifying a breakpoint or nearby intervals for joining, (3) dividing or combining continuous value ranges based on specific criteria, and (4) stopping this process at definite value.

MVD and UCP are promising approaches that are not supervised and helpful to apply to various ML issues other than the classification under adverse circumstances. They generalized a subset of the top global discretizers based on a compromise between UCPD, FUSInter, Distance, MDLP, and Chi2 as the ranges and accuracy \cite{bib4}. The possibility of utilizing multivariate discretization features may be investigated since parallel computers are becoming strong. Chi2 may delete redundant features, and Contrast or ID3 (dynamic discretization methods) may be addressed to integrate discretization into a learning process \cite{liu2002}.

\subsection{Forecasting}
Forecasting can be called predicting or estimating a value from the future \cite{armstrong2001}. Forecasting in a business-like SC performance is vital for suppliers that do forecast more than those that do not \cite{forslund2007}. Mainly three types of forecasting are done based on the length of the forecast: operational forecasting for short-term operational activities that range from hours to a few weeks, tactical forecasting for a moderate duration to support tactical planning that ranges from months to a few years, and strategic forecasting which is aligned with long-term goals to make strategic decisions \cite{lapide1999}. Furthermore, the frequency of a type of forecasting that is done is dependent on the length of the forecast. Long-term forecasts are rarely done, whereas operational forecasts may be required frequently. The different forecasts deal with different uncertainties. Long-term forecasts deal with raw material cost fluctuations, final product price changes, seasonal variations in demand, and changes in production rate in the long term. In contrast, short-term uncertainties are concerned with variations in daily processes, order cancellations, random failures in production, etc. \cite{gupta2003}. 

While forecasting is practical, forecasting correctly with more accuracy is even more helpful. Demand forecasting that allows anticipating sales in the forecasted period helps minimize overproduction and overstock \cite{chambers1971}. 

\subsubsection{Types}
Although forecasting techniques have evolved, forecasting techniques may be divided into three main categories: qualitative techniques, which deal with qualitative data or information to forecast; time series analysis and projection, which are related to historical data and patterns arising from them; causal models, where along with the historical data, special events and their relation with system elements are also considered \cite{chambers1971}. The qualitative technique is not much related to BD and data analytics; the other two are. Even so, qualitative data can be used to adjust forecasting models toward incredible accuracy. \cite{kuo2016} displayed one such example: Qualitative data can be used through fuzzy NNs combined with quantitative data for training the model. Nevertheless, accurate forecasts cannot be done based on qualitative data only. Time-series analysis is pretty straightforward, especially with the recent advancement of statistical tools. However, the role of such forecasts is to reduce errors in the forecast by minimizing the deviations at each point. Therefore, they do not consider special occasions such as promotions where sales are more remarkable than usual \cite{debaets2018}. This flaw brings us to causal models or models that consist of probabilities of forecasting accuracy, the effect of outside interventions, and the interrelation of different types of variables in the model \cite{hitchcock2023}. 

With the evolution of knowledge, different techniques for forecasting have emerged, and new classifications to understand them. \cite{buchatskaya2015} classified the different techniques into two broad groups of Intuitive and Formalized methods and divided Formalized methods further into Mathematical, System-structural, Associated, and Advanced information methods. 

\subsubsection{Model fit and train}
The dataset can be randomly split into the train, validation, and test sets for unbiased evaluation with new data to evaluate predictive performance with data different from training data. The best approach would be to split a dataset by a date feature. The most recent samples can be utilized for validation and testing. The primary concept is to choose a sample subset that accurately reflects the model data. 

Two factors determine the proportions of these three sets: the number of data samples and training models. Some models require significant training data; therefore, the model should be tuned for more extensive training sets in this scenario. Models with fewer hyperparameters will be easier to validate and tune, allowing a small validation set size. An extensive validation set will benefit if the model contains more significant hyperparameters. There will be no requirement for a validation set if the model has no hyper-parameters or is challenging to adjust.

When using k-fold CV, the train-test dataset splitting is repeated for k-times, with each new set being given a shot at becoming the hold-out set. Time-series data cannot be used with k-fold CV directly since they believe there is no connection between the rows and that they are all separate instances. For time-series data, instances' time horizon prevents arbitrarily dividing them into clusters. Instead, data should be segmented, and the chronological sequence of instances maintained. The term backtesting is used in time-series forecasting to describe the technique of evaluating models using past data. In meteorology, this is regarded as 'hindcasting' rather than 'forecasting.'

\subsubsection{Hyperparameter tuning}
Optimizing performance requires tuning hyperparameters automatically by Automated ML (AutoML). Hyperparameters are available in most ML systems. Hyperparameter adjustment has the most influence in optimizing, regularising, and architecting NNs. Common use cases of Automatic Hyperparameter Optimization (HPO) include:
\begin{itemize}
    \item ML, specially AutoML, will require less manual effort
    \item ML algorithms' efficiency (by customizing them to the task at hand) has improved, resulting in the new high state-of-the-art for significant ML standards in research findings \cite{snoek2012}
    \item It enhances the opportunity to reproduce the ML process
    \item It allows the fair comparison of methods with the same type of tuning.
\end{itemize}
One issue with HPO is that a particular configuration does not work well for all datasets \cite{kohavi1995}. These days, optimizing hyperparameters above the default parameters supplied by standard ML packages is increasingly acknowledged.

To assess a lower-cost optimization model, the authors proposed Bayesian Optimization and Hyperband (BOHB) as an efficient, flexible, stable, and parallelizable default HPO technique \cite{feurer2019}. However, if all hyperparameters are valid and just a few function evaluations are available, the (Spearmint) Gaussian technique is recommended \cite{snoek2012}. To solve restricted optimization issues in vast areas, they suggested RandomForest-based Tree Parzen Estimator (TPE) or Sequential Model-based Algorithm Configuration (SMAC) and Covariance Matrix Adaptation - Evolution Strategy (CMA-ES). Genetic approaches were initially used for adjusting two hyperparameters of RBF-SVM $C$ and $\mu$ faster than GridSearch for better forecasting accuracy \cite{chen2004}. CMA-ES was initially utilized for the optimization of hyperparameters to optimize hyperparameters of SVM $C$ and $\alpha$, (for all input sizes) the kernel scale of length $l_i$, and the whole matrix of spin and scaling \cite{friedrichs2005}. CMA-ES has lately proved a perfect solution for Parallelized HPO, superior to current Bayesian heuristics while optimizing 19 deep-NN hyperparameters on parallel 30 GPUs \cite{loshchilov2016}. A Gaussian online approach incorporated EI to tune the SVM hyperparameters, attaining factor 100 (regression, three hyperparameters) and 10 (classification, two hyperparameters) speedups against GridSearch \cite{frohlich2005}.  A robust, adaptable, and analogous combination of Hyperband and Bayesian optimization was introduced that significantly surpassed both BlackBox and Hyperband optimization for a broad variety of issues, along with SVM adjustment, different types of NNs, and reinforced ML algorithms \cite{falkner2018}. As early as 2002, ancient ML models offered GridSearch for hyperparameter optimization \cite{john1994, michie1995}. PatternSearch and GDFS (Greedy Depth-First Search) were the first dynamic optimization techniques for HPO, with GDFS outperforming GridSearch. Particle Swarm Model Selection (PSMS) handles conditional configuration space with a customized particle swarm optimizer. Modified Ensembling was added to PSMS to prevent overfitting and integrate the better methods from many generations \cite{escalante2010}. In addition, to maximize pipeline architecture and solely utilize Particle Swarm Optimization for every pipeline hyperparameter, PSMS was modified to utilize a genetic optimization algorithm \cite{sun2013}. For the hyperparameter adjustment of deep neural, \cite{bergstra2011} utilized Bayesian optimization, outperforming random searching and manual. In addition, TPE generated better output than a Gaussian approach considering the mechanism. Random forest TPE and Bayesian optimization have also succeeded in searching for combined neural and HPO \cite{bergstra2013}. 
We suggest a unique manual approach that might be helpful in general cases:
\begin{itemize}
    \item If there are many hyperparameters, the CV score can be evaluated for the first hyperparameter. After that, such a hyperparameter value should be selected to avoid overfitting and lower accuracy. After setting that hyperparameter, the next hyperparameter should be evaluated by iterating a similar process one by one. The HPO algorithm should be chosen based on the hyperparameter type. 
    \item If there are less than or equal to two hyperparameters, the desired HPO approach can be used directly.
\end{itemize}

\subsubsection{Model evaluation}
In our framework, we suggest model fitting and training on our analysis-ready training data using default parameters and then move to the next step of tuning the hyperparameters. If the model accuracy deteriorates, it is not the feature's fault; instead, we should focus on the HPO of the models. After HPO, the top-performing models can be easily chosen based on the elimination process, but the model-overfitting issue should be considered. The model can be evaluated with the predicted sales against the actual sales after at least one month in the initial operating period. 

\subsubsection{Top forecasting models}
Table \ref{forecasting_models} provides a list of time-series demand forecasting models that have been used in our reviewed literature. Table \ref{ml--models} provides a comprehensive overview of the most recently proposed ML models in different forecasting applications and the corresponding performance metrics evaluated in each literature. Considering accuracy and precision in forecasting the future time-series lags, the ARIMA model outperformed the AR (AutoRegressive), MA (moving average), and SES (Simple Exponential Smoothing) models. The empirical research reported that long short-term memory (LSTM) enhanced forecasting by 85\% when evaluated by comparison to ARIMA (traditional-based model). Furthermore, the number of epochs (training times) did not influence the forecasting model’s performance, which showed genuinely random behavior \cite{siami2018}.

\begin{table*}[!ht]
  \centering
  \caption{List of Time-Series Demand Forecasting Models Used in Literature Review}
  \label{forecasting_models}
  \begin{tabular}{|l|l|}
    \hline
    \textbf{Model Name} & \textbf{Citation} \\
    \hline
    ARIMA & \cite{yue2007} \\
    \hline
    LSTM & \cite{yue2007}, \cite{li2023} \\
    \hline
    RBFNN & \cite{yue2007} \\
    \hline
    Winter models with SVM & \cite{yue2007} \\
    \hline
    Adaptive Network & \cite{efendigil2009} \\
    \hline
    Fuzzy reasoning strategy and ANN & \cite{efendigil2009} \\
    \hline
    GAD & \cite{li2018} \\
    \hline
    SHEnSVM & \cite{yue2010} \\
    \hline
    Deep--Learning (DL) technique & \cite{kilimci2019} \\
    \hline
    SVR & \cite{ribeiro2022} \\
    \hline
    RF & \cite{ribeiro2022}, \cite{mitra2022} \\
    \hline
    XGBoost & \cite{gurnani2017}, \cite{ribeiro2022} \\
    \hline
    Ada Boost & \cite{shukla2022}, \cite{mitra2022} \\
    \hline
    Random Forest & \cite{shukla2022}, \cite{mitra2022}, \cite{hamdan2023} \\
    \hline
    MLP & \cite{shukla2022}, \cite{zohdi2022} \\
    \hline
    CNN--LSTM & \cite{nithin2022}, \cite{sandhya2022} \\
    \hline
    GRU & \cite{hu2023} \\
    \hline
    EGD--SNet & \cite{mehmood2022} \\
    \hline
    LSTM, BiLSTM, GRU & \cite{bassiouni2023} \\
    \hline
    Temporal Convolutional Network & \cite{bassiouni2023} \\
    \hline
    Swish Activation & \cite{nithin2022} \\
    \hline
    Shallow NN & \cite{rathipriya2023} \\
    \hline
    DNN & \cite{rathipriya2023} \\
    \hline
    Extreme Learning Machine (ELM) & \cite{chaudhuri2022} \\
    \hline
    Adaptive Neuro-Fuzzy Inference System (ANFIS) & \cite{hamdan2023} \\
    \hline
    SARIMAX & \cite{hamdan2023} \\
    \hline
    Prophet & \cite{hamdan2023}, \cite{mitra2022} \\
    \hline
    RF--XGBoost--LR & \cite{mitra2022} \\
    \hline
    RNN & \cite{sandhya2022} \\
    \hline
    GRU & \cite{sandhya2022} \\
    \hline
    XGBoost--LSTM & \cite{wei2021} \\
    \hline
    FB--Prophet & \cite{jha2021} \\
    \hline
    XGBoost--LightGBM & \cite{yang2021} \\
    \hline
    M--GAN--XGBoost & \cite{wang2021} \\
    \hline
    AUG--NN & \cite{javeri2021} \\
    \hline
  \end{tabular}
\end{table*}

For the demand forecasting procedure, \cite{yue2007} evaluated statistical models, RBFNN (Radial Basis Function NNs), and winter models with SVM. According to their conclusion, the efficiency of SVM outperforms other algorithms by about a mean MAPE outcomes threshold of 7.7\%. \cite{efendigil2009} demonstrated a new AI-utilized forecasting approach evaluating a fuzzy reasoning strategy and ANN based on the Adaptive Network to handle the demand containing inadequate knowledge. During testing, they obtained MAPE 18\% on average for some products. For unpredictable customer demands, neural methods have provided a robust forecasting strategy in a multi-level SC framework. A Greedy Aggregation Decomposition (GAD) approach is a generic approach to self-development in a discontinuous time-series forecasting method that considers double-based causes of variation, addressing a practical discontinuous issue of forecasting demand \cite{li2018}. With a limited dataset, they outperformed SBA, Croston’s method, TSB, MA-7, SES, MAPA, MA-3, ADIDA, iADIDA, and N-7 with a MAPE accuracy rate of 5.9\%. \cite{yue2010} offered the SHEnSVM (Selective \& Heterogeneous Ensemble of SVMs) model for sales forecasts. Individual SVMs were trained using samples produced by the bootstrap method, and grid search created parameters, as stated by the specified model. The optimum specific combo strategy was found using a genetic algorithm. They claimed a 10\% increment using the SVM algorithm \& a 64\% on average enhancement in MAPE. The authors used beer data from three product variants in their tests. \cite{kilimci2019} integrated DL technique, the SVR algorithm, and best-performing time-series analytic models utilizing the boosting ensemble approach to demand forecasting systems. Their DL implementation in the new integration strategy (MAPE: 24.7\%) lowers mean forecasting error in the SC, outperforming both the conventional best-performing forecasting model (MAPE: 42.4\%) and the unique integration strategy without DL (MAPE: 25.8\%). XGBoost, ARIMA, and Snaive STL decomposition have outperformed solo and hybrid models and the modeling mix and have provided the best forecasting accuracy \cite{gurnani2017}.

\cite{li2023} used Facebook Prophet (FB-Prophet) and ANNs to forecast lithium mineral resource prices in China. Quality and quantity of lithium data, network architecture, and activation functions significantly impacted the performance of an ANN forecasting model. Overfitting can occur when an ANN model is too closely tailored to the training dataset, and regularization and early halting strategies can enhance the model's performance. The FB-Prophet model, which uses a decomposable time-series model, can effectively forecast data with fewer value matrices, handle missing values, and practice adjustments. \cite{hu2023} created recurrent NN (RNN), LSTM, and gated recurrent unit (GRU) models to forecast the demand for U.S. influenza vaccinations, with data from 1980 to 2011 serving as the training set and data from 2012 to 2020 serving as the testing set. The prediction models may be scalable because there was no overfitting between the expected and actual numbers. The error comparisons demonstrated that GRU is more precise than LSTM and RNN in predicting vaccination demand. Energy Generation and Demand Forecasting Search Net (EGD-SNet), a framework that can anticipate energy production, demand, and temperature across various areas, was reported in the study of \cite{mehmood2022}. Together with the 

The 10 most popular ML regressors in the EGD-SNet framework include 11 dimensionality reduction techniques and 13 alternative feature selection algorithms. It employs a Particle Swarm Optimizer (PSO) to train regressors intelligently by locating the best hyperparameters. Also, it can create an end-to-end pipeline by selecting the right regressor, feature, and dimensionality reduction methodologies to accurately anticipate energy generation or demand for a specific geographical data set, depending on the features of the data. \cite{bassiouni2023} implemented many DL methods, including data collecting, de-noising or pre-processing, feature extraction, and classification stages. Two primary DL models determine feature extraction. The first variation used three RNN structures: LSTM, BiLSTM, and GRU. The second variant used the temporal convolutional network (TCN). They used SoftMax, RT, RF, KNN, ANN, and SVM classifiers for an online dataset. TCN predicts COVID-19-restricted shipping risk almost 100\% accurately.

As the daily fish demand forecasting models for grocery merchants to reduce food waste and enhance sustainable SCs, \cite{migueis2022} investigated LSTM, Feedforward NNs, Support Vector Regression, RF, and a Holt-Winters statistical model. The findings showed that the LSTM model provided the best outcomes in terms of root mean squared error (27.82), mean absolute error (20.63), and mean positive error (17.86). \cite{haider2022} forecasted solar Global Horizontal Irradiance using statistical and Deep Learning architectures, which aids grid management and power distribution and highlights Pakistan's solar power potential in addressing global climate change. They employed SARIMAX, Prophet, LSTM, convolutional NN (CNN), and ANN statistical approaches. Error measures like $R^2$, MAE, MSE, and RMSE were used to evaluate each model's performance. They concluded that SARIMAX and Prophet are ideal for long-term forecasts, whereas ANN, CNN, and LSTM are best for short-term forecasts. \cite{shukla2022} found that the optimum model for every participant in the SC across all three inventory replenishment strategies is a stacked ensemble model consisting of XG Boost, Ada Boost, and Random Forest. According to a methodology for comparing forecasting methods developed by \cite{singha2022}, the MLP method has a little edge over the CNN, LSTM, and CNN-LSTM approaches. \cite{vithitsoontorn2022} utilized data from the last five years to estimate demand for eight dairy products from five dairy production facilities using a direct multistep prediction method. ARIMA works effectively on a narrow subset of unpredictable series, whereas LSTMs excel at anticipating seasonal patterns. It outperforms ARIMA for trends. Monthly data decreased model training error.

For the purpose of forecasting daily energy use, \cite{ribeiro2022} investigated the effectiveness of three ML models, SVR, RF, and XGBoost; three deep learning models, RNNs, LSTM, and GRU; and ARIMA. For both very short-term load forecasting (VSTLF) and short-term load forecasting (STLF), the suggested XGBoost models beat competing models; the ARIMA model did the worst. \cite{elmir2023} presented a smart platform for data-driven blood bank management that forecasts blood demand and balances blood collection and distribution based on optimal blood inventory management to avoid blood wastage and shortage. This improves blood quality and quantity, increasing blood collection by 11\% and reducing blood waste by 20\%. Balancing blood collection and distribution based on good blood inventory management and arranging blood donation sessions to avoid cancellations may lower inventory levels. \cite{nithin2022} proposed a CNN-LSTM model with Swish Activation to estimate a store's supply based on prior sale. This outperforms Rectified Linear Unit (ReLU), the most effective activation function. They forecasted sales using Multilayer Perceptron, LSTM cells, and CNNs. CNN-LSTM Model has a reduced RMSE, according to the experiment. Pharmaceutical businesses can use Shallow NN and DNN demand forecasting models for eight anatomical treatment chemical thematic drug groups \cite{rathipriya2023}. Shallow NN models performed well for five of eight medication categories, while the ARIMA model performed best for the other three.

\cite{chaudhuri2022} introduced an extreme learning machine (ELM) model using the Harris Hawks optimization (HHO) method to estimate e-commerce product demand. In forecasting product demand for the next three months, the ELM-HHO model outperformed the statistical ARIMA (7,1,0) model by 62.73\%, the NN-based GRU model by 40.73\%, the LSTM model by 34.05\%, the traditional non-optimized ELM model with 100 hidden nodes by 27.16\%, and the ELM-BO model by 11.63\%. \cite{hamdan2023} developed a novel ML forecasting approach by merging adaptive neuro-fuzzy inference system (ANFIS) and time-series data features to forecast real-time e-order arrivals in distribution hubs, helping third-party logistics providers better manage hourly-based e-order arrival rates. ELM, GB, KNN, MLP, and DT were five ML algorithms used by \cite{zohdi2022} to forecast demand in a business based on Black Friday customer information. According to the results, MLP, ELM, GB, KNN, and DT were the top algorithms in terms of MSE, while ELM, MLP, GB, DT, and KNN had the greatest performances in terms of MAE. Moreover, ELM had a higher $R^2$ value of 0.6365, whereas DT had a lower value (0.4877). \cite{mitra2022} compared RF, XGBoost, gradient boosting, AdaBoost, and ANN algorithms to a hybrid (RF-XGBoost-LR) model for retail chain sales forecasting. A US retail company's weekly sales data was used to analyze estimates based on factors like temperature and shop size. The hybrid RF-XGBoost-LR outperformed other models in many criteria. RNN and LSTM were used by \cite{sandhya2022} to improve stock price prediction. The memory cell, a computer that replaces artificial neurons, is buried in the network. The study increased epochs and load sizes to improve precision. Time and lot size boost prediction accuracy in this work. The test data predicts the specified technique, which yields more accurate outcomes. The proposed method forecasts stock markets more accurately.

\begin{table*}[!ht]

\caption{The Most Recent (2022--2023) ML Models for Forecasting Applications in SCM}
\label{ml--models}
\centering
\begin{tabular}{|p{2cm}|p{4cm}|p{3cm}|p{4cm}|}
\hline
\textbf{Study} & \textbf{ML Models} & \textbf{Performance Metrics} & \textbf{Application/Domain} \\
\hline
\cite{bassiouni2023} & EGD-SNet Framework: 10 ML regressors & Accuracy, RMSE, MAE & Energy generation and demand forecasting \\
\hline
\cite{migueis2022} & LSTM, Feedforward NNs, Support Vector Regression, RF, Holt-Winters model & RMSE, MAE, Mean positive error & Daily fish demand forecasting for grocery merchants \\
\hline
\cite{haider2022} & SARIMAX, Prophet, LSTM, CNN, ANN & $R^2$, MAE, MSE, RMSE & Solar Global Horizontal Irradiance forecasting \\
\hline
\cite{shukla2022} & Stacked ensemble model (XG Boost, Ada Boost, Random Forest) & Not specified & SC inventory replenishment \\
\hline
\cite{singha2022} & MLP, CNN, LSTM, CNN--LSTM & Not specified & Forecasting method comparison \\
\hline
\cite{vithitsoontorn2022} & ARIMA, LSTM & Not specified & Dairy product demand forecasting \\
\hline
\cite{ribeiro2022} & SVR, RF, XGBoost, RNNs, LSTM, GRU, ARIMA & Not specified & Daily energy use forecasting \\
\hline
\cite{elmir2023} & Smart platform utilizing ML models & Not specified & Blood bank management and demand forecasting \\
\hline
\cite{nithin2022} & CNN-LSTM model with Swish Activation & RMSE & Store supply forecasting based on prior sale \\
\hline
\cite{rathipriya2023} & Shallow NN, DNN, ARIMA & Not specified & Pharmaceutical demand forecasting \\
\hline
\cite{chaudhuri2022} & Extreme Learning Machine (ELM) with Harris Hawks optimization (HHO) & Not specified & E-commerce product demand forecasting \\
\hline
\cite{hamdan2023} & Adaptive Neuro-Fuzzy Inference System (ANFIS) & Not specified & Real-time e-order arrivals forecasting \\
\hline
\cite{zohdi2022} & ELM, GB, KNN, MLP, DT & MSE, MAE, $R^2$& Business demand forecasting based on Black Friday customer information \\
\hline
\cite{mitra2022} & RF, XGBoost, Gradient boosting, AdaBoost, ANN, Hybrid RF-XGBoost-LR & Not specified & Retail chain sales forecasting \\
\hline
\cite{sandhya2022} & RNN, LSTM & Not specified & Stock price prediction \\
\hline
\end{tabular}

\end{table*}

Based on the mentioned studies, we suggest considering the following recently best-performing hybrid time-series demand forecasting ML models: 1. XGBoost-LSTM \cite{wei2021} 2. FB-Prophet \cite{jha2021} 3. XGBoost-LightGBM \cite{yang2021} 4. M-GAN-XGBoost \cite{wang2021} 5. SARIMA integrated AttConvLSTM, and FB-Prophet \cite{wan2021} 6. AUG-NN \cite{javeri2021}.

When selecting the primary top-forecasting model, it is recommended to consider the best cross-validation (CV) score, minimum runtime, and space consumption as criteria for evaluation.

\section{Control-process}
\label{sec4}
Actual processes do not always go as predicted. There are variabilities in performances emerging from changing levels of efficiency. In ideal cases, the workforce and existing capacity can achieve the goal as planned. However, performance levels are inconsistent with humans \cite{rabbitt2001} and vary with other factors such as WIP inventory, machine utilization, product mix, and queueing system \cite{wu2005}.  Such fluctuations in efficiency cannot be predicted accurately. Hence, they must be recognized in time, and appropriate measures to meet the requirements must be taken. Responsive to the SC process's randomness may consist of three steps: capturing or recording data simultaneously with the SC activity, comparing the recorded data with the standard, and adjusting capacity to meet the short-term goal. Data can be used to determine the optimal decision for the changing suppliers, changing price levels, the competitiveness of the competitors, and monitoring performance during the process \cite{elgendy2014}. Furthermore, during a discrepancy in planned and actual output levels, the root cause can be identified using BDA \cite{russom2011}. \cite{bagshaw2017} mentioned other benefits of BD on workforce scheduling, production efficiency, employee productivity, capacity utilization, flexibility, and lead time reduction.

\subsection{Information flow}
Forecasting decisions affect further SC planning. As such, information flows across multiple phases of the SC process. Logistics superiority and better stock level synchronization are possible through a flow of demand information from downstream members to the upstream ones and the flow of production plan and delivery information from the upstream members to the downstream ones \cite{vanpoucke2009}. Like the three types of forecasting, there are three types of decisions in SC: strategic, operational, and tactical \cite{strack2010}. Recent studies have shown how the findings from one level can affect other decisions and limit the number of options for subsequent decisions \cite{berg1999}.

It is possible to attain efficiency through forecasting by properly allocating resources in different areas such as workforce, capacity, inventory management, etc. There are need-based, supply-based, and demand-based models for forecasting and planning in such areas of SCM \cite{safarishahrbijari2018}. Each method requires some sort of information flow. The forecasted amount can be used to estimate the number of dependent inventory demands. The final product’s demand indirectly affects the required workforce, capacity, warehouse planning, and lower costs through optimization.

\subsection{Production efficiency}
Having real-time data on production boosts production efficiency. Firms can manage order processing across SCs and companies while decreasing errors and waste inside manufacturing facilities by incorporating real-time data into SC operations \cite{waller2013}. This efficiency is further enhanced when data from suppliers and distributors are available. Through close connections and sharing information with SC partners, data-driven SCs may also affect manufacturing and operations processes through increased efficiency in product development, product design, quality improvement, and balance between capacity and demand \cite{sanders2016}. Additionally, data integration in the SC has been found to aid in developing production strategies and the timely delivery of products and services \cite{droge2012}.

\subsection{Employee productivity}
In general, there is either an excess of the workforce or a shortage in the production process; the question is how to reduce the inefficiency. Under variable output requirements, workforce scheduling without data analysis entails investing in cross-job training to enable workers to be more productive and efficient in their work. However, this reduces performance as time is spent on upskilling or reskilling.

Data-driven decision-making, or the forecasting of required outputs to estimate the required workforce, is an excellent way to minimize such costs of hiring and laying off by adequately scheduling the workforce \cite{bagshaw2017}. Through proper scheduling, the workforce from idle time can be shifted for workdays requiring extra hours and thereby balanced. \cite{noack2008} showed that work pressure could be balanced with reduced slack time and workforce through different heuristic algorithms, with each algorithm performing well in different areas of efficiency. 

\subsection{Inventory management}
Reducing costs from inventory can cut the overall cost of the business. Different models have been created to minimize costs and maximize profits that aid with material planning mechanisms, stock-out predictions, inventory level predictions, and many more \cite{hajek2020}. Inventory costs can be lowered at sourcing, transportation, and holding levels, optimizing inventory decisions \cite{sanders2016}.

\subsection{Role of data by time frame}
Although BDA can make SC processes efficient, it cannot be done with the same forecast. Capacity planning or storage size falls under long-term strategic decisions requiring long-term forecasts or aggregated short-term forecasts. Contrarily, production plans may be short-term operational decisions requiring short-term predictions. \cite{andrawis2011} stated how predictions could be derived from the aggregation of short-term, disaggregation of long-term, or a co-integration of both kinds of forecasts. Hence, on the one hand, separate forecasts can be produced. Conversely, forecasts can be derived from other forecasts to maintain relevance.

\section{Post-process}
\label{sec5}
\subsection{SC performance}
Once an SC process is completed, the performance needs to be reviewed to identify the gaps in planning models. Performance measurement is defined as quantifying actions across two fundamental dimensions: effectiveness and efficiency \cite{neely1995}. Performance measurement is essential to control the output; without it, no person or machine can be held liable for subpar performance, and problems will be harder to identify and solve. Performance measurement helps with information for management feedback, decision-making, monitoring performance, diagnosing problems, motivating people, identifying potentials of a decision, measuring success or failure, reviewing and adjusting business strategies, specifying company goals, and much more \cite{chan2003}. \cite{ho2007} offered a comprehensive methodology considering all three SC system stages, including ERP-based SC performance. To comprehend whether network scanning and embeddedness are linked to SC performance, Bernardes and Zsidisin (2008, 209) studied the correlation between SCM strategy and network scanning and embeddedness concepts.

Immediately after the tasks are completed, the performance data must be recorded. For data collection, the performance metrics are first to be identified, just as those found for business evaluations \cite{bittencourt2005}. \cite{gunasekaran2001} mentioned plan success, source optimization, production efficiency, delivery performance, and customer support-relation and satisfaction, each having multiple performance metrics under them. The performance level found afterward can be of three types: below average, average, and above-average \cite{asrol2021}. The actions followed after such a finding are different in each case. When a below-average performance is observed, managers can either look for anomalies in the system or review whether the goals set were too high to achieve.

Conversely, an above-average performance requires rechecking the goals so that optimization of resources is possible. In order to meet the goals-setting theory stated by \cite{locke2006}, these changes may be adjusted to suit. Operational benefits such as performance monitoring, objective setting, management, transparency, and planning functions can be improved with the assistance of BDA and performance metrics derived from them through the use of predictive KPIs, dashboards, and scorecards by the SC operational managers within the organization \cite{elgendy2014}.

Besides managerial decision-making, the performance data are crucial to modifying existing forecasting models. Under a considerable deviation of performance, the data received from this level needs to be sent back to the forecasting stage for tweaking the forecasting model to higher perfection. The performance metrics can thus act as an indicator of forecasting model errors. 

\subsection{Forecasting error measurement}
Our proposed cyclic framework is evaluated against predicted sales when the actual sales data is available or using the hold-out set. Nevertheless, the hold-out set might not be perfect for real-world scenarios, so we encourage a cyclic and continuous development process from real-sales data evaluation insights. We encourage a cyclic and continuous development process from real-sales data evaluation insights. A few evaluation metrics can be used for the post-process evaluation to fine-tune the forecasting model in the preprocessing phase. Assume test data with m periods, $t=1,\ldots,m$. The difference between forecasted sales $f_t$ and actual sales $y_t$ at a period $t$ can be referred to as the forecasting error $e_t=y_t-f_t$.

\subsubsection{Mean absolute error}
\begin{equation}
\small
    MAE = \frac{1}{m} \sum_{t=1}^m |e_t | = mean(|e_t |)
\end{equation}

MAE is very straightforward and relatively simple to explain, and scale dependence is its disadvantage.

\subsubsection{Mean absolute percentage error}
\begin{equation}
\small
	 MAPE = \frac{1}{m} \sum_{t=1}^m|\frac{e_t}{y_t}\times100\\%
\end{equation}

MAPE is perhaps the most often utilized error indicator for business forecasting because of its comprehensiveness. However, despite the term ‘Percentage,’ the MAPE value may be higher than 100\%. The rows equal to 0 causes problems since the fraction’s denominator cannot be filled in. MAPE is an appropriate metric when dealing with intermittent demand. Asymmetry is its major drawback as it penalizes overfitting more than underfitting, leading to probable skewness.

\subsubsection{Mean squared error}
\begin{equation}
\small
	MSE = \frac{1}{m} \sum_{t=1}^m {e_t}^2
\end{equation}

Compared to RMSE, MSE takes less runtime and is more flexible. However, we might not interpret MSE as the actual sales because the error is squared.

\subsubsection{Root mean squared error}
\begin{equation}
\small
	 RMSE=\sqrt{\frac{1}{m} \sum_{t=1}^m{e_t}^2} =\sqrt{mean(|{e_t}^2|)}
\end{equation}

Two consequences occur by performing the dual transformation in RMSE: more weight is placed on more significant errors, and positive and negative errors cannot cancel one another out since they are all transformed into positives.

\subsubsection{Mean absolute scaled error}
For non-seasonal time-series,
\begin{equation}
\small
	MASE=\frac{\frac{1}{J} \sum_{j}|e_j|}{\frac{1}{T-1}\sum_{t=2}^T|y_t-y_(t-1)|}
\end{equation}
For seasonal time-series,
\begin{equation}
\small
	MASE=\frac{\frac{1}{J} \sum_{j}|e_j|}{\frac{1}{T-m}\sum_{t=m+1}^T|y_t-y_(t-m)|}
\end{equation}

With MAE, outliers are protected; with RMSE, we are assured of an impartial prediction. SC Analysts need to analyze MAE and see whether it results in a significant bias; therefore, they should utilize RMSE. In situations when there are many outliers in the dataset, MAE may help correct the skewed prediction.

\subsubsection{Tracking signal}
The tracking signal is the way to verify if the current forecasting method is correct. A tracking signal that changes according to the forecast bias shows bias in the prediction model. It is often employed when the forecasting model’s validity is questionable.
\begin{equation}
\small
	\text{Algebraic sum of forecast error} = \sum_{t=1}^m|e_t|
\end{equation}
\begin{equation}
	\text{Tracking Signal} = \frac{\text{Algebraic sum of forecast error}}{\text{Mean Absolute Error}}
\end{equation}

A rule of thumb holds that the technique employed for forecasting is accurate when the tracking signal is within $-4$ to $+4$.

\subsection{Phantom inventory}
Research has shown that imprecise perpetual inventories (PIs) are overestimated approximately 50\% of the time; that is, PI displays higher stock than that is present in the shop, called a phantom inventory. The most severe issue in a phantom inventory is unavailability – the system considers it has an adequate inventory and does not order a replenishment. The recognized reasons for phantom inventory are \cite{waller2006}:
\begin{itemize}
    \item Stolen goods, defective products that are not reported
    \item Cashier mistakes
    \item A shop may get deliveries from the distribution center (products that should have but were not received)
    \item Returned goods that should update the system are sometimes wrong.
\end{itemize}

To resolve the stock inconsistency, businesses may perform a bunch of tasks \cite{kang2005}:
\begin{itemize}
    \item The supply of safety may be raised. The enhanced security inventory aims to mitigate inventory problems by having ‘excess’ inventories at hand. RFID may reduce the costs of storing this additional and redundant inventory.
    \item The business may often conduct manual inventory numbers. Physical inventory audits may interrupt storage, are extremely expensive, and differ in precision – improved RFID precision may be an affordable option.
    \item The business may construct a continuous decrease equivalent to the total inventory loss that one believes takes place to balance the phantom inventory. The issue is that the precise inventory loss is not known. The visibility provided by RFID may be more accurate than conventional stock loss techniques.
    \item The business may attempt to minimize mistakes by improving inventory management, decreasing fraud, etc. 
\end{itemize}

Inventory precision determines forecasting, procurement, and replenishment quality, where inventory records are used as input. Inaccurate demand forecasting due to phantom inventory (overstated PI) may be improved by including RFID in the process \cite{hardgrave2009}.

\section{Challenges}
\label{sec6}
This section aims to provide a comprehensive overview of the challenges encountered during the review of 152 articles from 1969 to 2023 in the field of BDA-SCM for forecasting, with a specific focus on data preprocessing and ML techniques. The challenges identified herein will serve as a valuable resource for future researchers, enabling them to address and overcome these obstacles, ultimately advancing the domain and contributing to its growth and development.
\subsection{Data quality and reliability}
One of the critical challenges observed in the reviewed literature is the issue of data quality and reliability. Many studies acknowledged the presence of incomplete, inconsistent, and erroneous data within SC datasets. Future research efforts should focus on developing robust data cleansing, integration, and quality assurance techniques to enhance the reliability and accuracy of the forecasting models.
\subsection{Scalability and performance}
With the exponential growth of data in SCM, scalability and performance have become significant challenges. The reviewed articles often lacked details on how their proposed techniques would scale up to handle large-scale datasets or real-time processing requirements. Future researchers should explore scalable algorithms, distributed computing frameworks, and parallel processing techniques to ensure the effectiveness and efficiency of forecasting models.
\subsection{Variety and complexity of data Sources}
The diverse range of data sources, such as structured, unstructured, and semi-structured data, presents challenges in data preprocessing and feature extraction. The reviewed literature indicated limited exploration of techniques for effectively handling data variety and complexity. Future research should focus on developing innovative methods for integrating and analyzing heterogeneous data sources to extract meaningful insights for accurate forecasting.
\subsection{Feature Engineering and Selection}
Effective FE and to improve forecasting accuracy, including automated FS, dimensionality reduction, and feature representation approaches to identify the most relevant features for forecasting within the SC context. Future researchers should investigate advanced FE techniques, including automated FS, dimensionality reduction, and feature representation, to improve forecasting accuracy.
\subsection{Model interpretability and explainability}
The black-box nature of some ML models limits their interpretability and hampers decision-making processes. The surveyed literature revealed a lack of emphasis on model interpretability, hindering the wider adoption of forecasting techniques in SCM. Future research should focus on developing transparent and interpretable models that provide explanations for their predictions, enabling practitioners to understand and trust the results.
\subsection{Real-time data processing and analysis}
SCM requires real-time monitoring and decision-making capabilities. However, the surveyed literature demonstrated a limited exploration of real-time data processing and analysis techniques for forecasting purposes. Future research efforts should concentrate on developing real-time forecasting frameworks that leverage stream processing, online learning, and adaptive algorithms to handle dynamic and time-sensitive SC scenarios.
\subsection{Privacy and security concerns}
Integrating big data in SCM raises concerns regarding data privacy and security. The surveyed articles paid limited attention to these challenges, and there is a lack of comprehensive approaches to ensure the privacy and security of sensitive SC data. Future researchers should focus on developing robust privacy-preserving and secure ML techniques to safeguard data while maintaining the accuracy and efficiency of forecasting models.
\subsection{Integration of domain knowledge}
SCM involves complex domain-specific knowledge, including industry-specific constraints, regulations, and contextual factors. The reviewed literature showed a limited integration of such domain knowledge into the forecasting frameworks. Future research should emphasize the incorporation of domain expertise and contextual information to enhance the relevance and accuracy of forecasting models within the SC domain.
\subsection{Lack of benchmark datasets and evaluation metrics}
The absence of standardized benchmark datasets and evaluation metrics hinders the comparison and reproducibility of forecasting techniques. The reviewed articles often utilized different datasets and evaluation metrics, making it challenging to assess the performance of various models. Future researchers should strive to establish benchmark datasets and evaluation protocols specific to SC forecasting, enabling fair comparisons and facilitating advancements in the field.

By overcoming these challenges through innovative techniques and methodologies, researchers can contribute to the advancement of this field, leading to more accurate, scalable, and interpretable forecasting models for SCM.

\section{Practical implications}
\label{sec7}
The findings of this research offer substantial practical implications for SC practitioners, providing actionable insights that can be effectively implemented in real-world scenarios. The proposed BDA-SCM framework serves as a strategic guide, and its practical application holds the potential for significant benefits in enhancing overall SC operations.

SC practitioners can implement the BDA-SCM framework by initially aligning data collection methodologies with specific SC objectives. This involves adopting a systematic approach to gathering data directly relevant to the SC ecosystem's unique dynamics and challenges. By integrating the framework into their operational processes, practitioners can leverage the power of BDA at various stages, from problem identification to performance evaluation.

The implementation of the BDA-SCM framework promises several tangible benefits for SC practitioners. Firstly, the framework enhances the accuracy of forecasting models, providing practitioners with more reliable insights into demand patterns, inventory needs, and workforce requirements. This, in turn, enables optimized decision-making across various facets of SCM. Secondly, the cyclic connection within the framework ensures adaptability to dynamic SC conditions. SC practitioners can continuously refine and optimize their forecasting models based on real-time data, thereby staying responsive to changing market dynamics and mitigating potential disruptions. Furthermore, the framework's emphasis on KPIs and error-measurement systems enables practitioners to evaluate and improve the performance of their forecasting models systematically. This enhances operational transparency and contributes to the overall efficiency and planning effectiveness of the SC.

In practical terms, the BDA-SCM framework supports inventory management by providing accurate demand forecasts, aids in determining workforce needs, optimizes cost factors, and facilitates efficient production and capacity planning. By fostering a holistic approach to SCM, the framework equips practitioners with a systematic and data-driven strategy to address the intricacies of modern SC dynamics. In essence, the practical implementation of the BDA-SCM framework empowers SC practitioners to navigate the complexities of their operational environments with greater precision and foresight, ultimately contributing to enhanced resilience, efficiency, and competitiveness in the ever-evolving SCM landscape.

\section{Conclusions}
\label{sec8}
This systematic review diligently identified and compared state-of-the-art SC forecasting strategies and technologies within the defined temporal scope, conducting a comprehensive review of 152 papers spanning from 1969 to 2023. This study has made significant strides in addressing the challenges inherent in SC forecasting, offering cutting-edge technological solutions within the context of a comprehensive BDA-SCM framework. The key findings and contributions of this study can be summarized as follows:
\begin{enumerate}
    \item Pre-process: In the pre-processing stage of SC forecasting, the significance of accurate data aligned with SC objectives was emphasized. The study provided recommendations for SC analysts, including using EDA, FE, hyperparameter tuning, and recent ML model training approaches to improve forecasting accuracy. However, it is essential to note that further research is needed to explore advanced techniques for data cleansing, integration, and quality assurance to ensure reliable and high-quality input data.
    \item Control-process: The study discussed how BD could facilitate efficient managerial decision-making in various areas of SCM, such as production \& capacity planning, workforce requirements, and inventory management. Leveraging insights from forecasted data allows decision-makers to optimize SC operations and resource allocation. However, future research should focus on developing real-time decision support systems that can integrate and analyze large-scale data streams to enable timely and effective decision-making.
    \item Post-process: The post-process section emphasized SC performance measurement and the role of BDA in optimizing model predictions. By analyzing performance metrics and leveraging BDA techniques, SC practitioners can identify areas for improvement and refine their forecasting models accordingly. Future research efforts should focus on developing comprehensive performance measurement frameworks specific to SC forecasting, including quantitative and qualitative metrics, to enable more accurate evaluation and comparison of forecasting models. Additionally, the study addresses the accuracy of inventory records as a crucial determinant for forecasting, procurement, and replenishment quality. Mitigating inaccuracies resulting from phantom inventory is highlighted, with the inclusion of RFID technology in inventory management processes as a viable solution. Future research should explore advanced techniques and methodologies to address phantom inventory, incorporating emerging technologies and developing comprehensive inventory management and forecasting frameworks to enhance overall SC performance.

\end{enumerate}

This study has successfully addressed the research questions posed:

RQ1: The study has identified and outlined efficient steps to formulate an ML forecasting model for predicting SC factors. Recommendations for accurate data preprocessing, FE, hyperparameter tuning, and advanced ML model training approaches have been provided to enhance the accuracy of SC forecasting models.

RQ2: The study has emphasized the importance of connecting, tracking, and optimizing the forecasting, SC decision-making, and performance measurement processes in a cyclic order. The proposed BDA-SCM framework encompasses the Pre-process, Control-process, and Post-process stages, providing guidance on integrating these processes to optimize SC operations and resource allocation.

RQ3: The study has explored the impact of forecasting on SC performance and identified relevant ML forecasting models for SC forecasting. The connection between accurate forecasting and improved SC performance has been highlighted, with recommendations for performance measurement and using BDA techniques to optimize model predictions.

By successfully addressing these RQs, this study contributes to the advancement of the field by providing insights into efficient ML modeling steps, the integration of forecasting and SC decision-making processes, and the relevance of ML forecasting models for SC forecasting. Future research should build upon these findings to further enhance the understanding and implementation of BDA in SCM.

While this systematic literature review (SLR) followed a rigorous and objective evaluation approach, acknowledging its limitations is crucial. These limitations include the availability of relevant literature, potential publication bias, and the dynamic nature of the BDA-SCM field. Future research endeavors should aim to address these gaps by conducting further empirical studies, developing benchmark datasets, and exploring emerging technologies and methodologies to advance the understanding and implementation of BDA in SCM.

By considering and addressing the challenges and limitations outlined in this study, future researchers can build upon its findings and contribute meaningfully to the continued advancement of the BDA-SCM domain.


\begin{thebibliography}{00}


\bibitem{hajek2020}
Hajek P, Abedin MZ A Profit Function-Maximizing Inventory Backorder Prediction System Using Big Data Analytics | IEEE Journals \& Magazine | IEEE Xplore. \href{https://ieeexplore.ieee.org/document/9046037}{ ieeexplore.ieee.org/document/9046037}. Accessed 23 Oct 2023

\bibitem{andrawis2011}
Andrawis RR, Atiya AF, El-Shishiny H (2011) Combination of long term and short term forecasts, with application to tourism demand forecasting. International Journal of Forecasting 27:870–886. \href{https://doi.org/10.1016/j.ijforecast.2010.05.019}{10.1016/j.ijforecast.2010.05.019}

\bibitem{arauzoazofra2011}
Arauzo-Azofra A, Aznarte JL, Benítez JM (2011) Empirical study of feature selection methods based on individual feature evaluation for classification problems. Expert Systems with Applications 38:8170–8177. \href{https://doi.org/10.1016/j.eswa.2010.12.160}{10.1016/j.eswa.2010.12.160}

\bibitem{armstrong2001}
Armstrong JS (2001) Principles of Forecasting: A Handbook for Researchers and Practitioners. Springer US, Boston, MA

\bibitem{asrol2021}
Asrol M, Marimin, Machfud, et al. (2021) Risk Management for Improving Supply Chain Performance of Sugarcane Agroindustry. Industrial Engineering \& Management Systems 20:9–26. \href{https://doi.org/10.7232/iems.2021.20.1.9}{10.7232/iems.2021.20.1.9}

\bibitem{bagshaw2017}
Bagshaw KB (2017) WORKFORCE BIG DATA ANALYTICS AND PRODUCTION EFFICIENCY: A Manager’s Guide. Archives of Business Research 5:. \href{https://doi.org/10.14738/abr.57.3168}{10.14738/abr.57.3168}

\bibitem{barnard1999}
Barnard J, Meng X-L (1999) Applications of multiple imputation in medical studies: from AIDS to NHANES. Stat Methods Med Res 8:17–36. \href{https://doi.org/10.1177/096228029900800103}{10.1177/096228029900800103}

\bibitem{bassiouni2023}
Bassiouni MM, Chakrabortty RK, Hussain OK, Rahman HF (2023) Advanced deep learning approaches to predict supply chain risks under COVID-19 restrictions. Expert Systems with Applications 211:118604. \href{https://doi.org/10.1016/j.eswa.2022.118604}{10.1016/j.eswa.2022.118604}

\bibitem{batista2003}
Batista GEAPA, Monard MC (2003) An analysis of four missing data treatment methods for supervised learning. Applied Artificial Intelligence 17:519–533. \href{https://doi.org/10.1080/713827181}{10.1080/713827181}

\bibitem{elmir2023}
Ben Elmir W, Hemmak A, Senouci B (2023) Smart Platform for Data Blood Bank Management: Forecasting Demand in Blood Supply Chain Using Machine Learning. Information 14:31. \href{https://doi.org/10.3390/info14010031}{10.3390/info14010031}

\bibitem{bergstra2011}
Bergstra J, Bardenet R, Bengio Y, Kégl B (2011) Algorithms for hyper-parameter optimization. Advances in neural information processing systems 24:

\bibitem{bergstra2013}
Bergstra J, Yamins D, Cox D (2013) Making a Science of Model Search: Hyperparameter Optimization in Hundreds of Dimensions for Vision Architectures. In: Proceedings of the 30th International Conference on Machine Learning. PMLR, pp 115–123

\bibitem{bernardes2008}
Bernardes ES, Zsidisin GA (2008) An examination of strategic supply management benefits and performance implications. Journal of Purchasing and Supply Management 14:209–219. \href{https://doi.org/10.1016/j.pursup.2008.06.004}{10.1016/j.pursup.2008.06.004}

\bibitem{bickel1981}
Bickel PJ, Doksum KA (1981) An Analysis of Transformations Revisited. Journal of the American Statistical Association 76:296–311. \href{https://doi.org/10.1080/01621459.1981.10477649}{10.1080/01621459.1981.10477649}

\bibitem{bittencourt2005}
Bittencourt F, Rabelo RJ (2005) A Systematic Approach for VE Partners Selection Using the SCOR Model and the AHP Method. In: Camarinha-Matos LM, Afsarmanesh H, Ortiz A (eds) Collaborative Networks and Their Breeding Environments. Springer US, Boston, MA, pp 99–108

\bibitem{bolon2013}
Bolón-Canedo V, Sánchez-Maroño N, Alonso-Betanzos A (2013) A review of feature selection methods on synthetic data. Knowl Inf Syst 34:483–519. \href{https://doi.org/10.1007/s10115-012-0487-8}{10.1007/s10115-012-0487-8}

\bibitem{brown2012}
Brown G, Pocock A, Zhao M-J, Luján M (2012) Conditional likelihood maximisation: a unifying framework for information theoretic feature selection. The journal of machine learning research 13:27–66

\bibitem{buchatskaya2015}
Buchatskaya V, Teploukhov PB and S (2015) Forecasting Methods Classification and its Applicability. INDJST 8:1–8. \href{https://doi.org/10.17485/ijst/2015/v8i30/84224}{10.17485/ijst/2015/v8i30/84224}

\bibitem{duarte2006}
Cardoso F, Duarte C (2006) The use of qualitative information for forecasting exports. Banco de Portugal Economic Bulletin, Winter 67–94

\bibitem{chambers1971}
Chambers JC, Mullick SK, Smith DD (1971) How to choose the right forecasting technique. Harvard University, Graduate School of Business Administration Cambridge, MA …

\bibitem{chan2003}
Chan FTS (2003) Performance Measurement in a Supply Chain. Int J Adv Manuf Technol 21:534–548. \href{https://doi.org/10.1007/s001700300063}{10.1007/s001700300063}

\bibitem{chaudhuri2022}
Chaudhuri KD, Alkan B (2022) A hybrid extreme learning machine model with harris hawks optimisation algorithm: an optimised model for product demand forecasting applications. Appl Intell 52:11489–11505. \href{https://doi.org/10.1007/s10489-022-03251-7}{ 10.1007/s10489-022-03251-7}

\bibitem{chen2004}
Chen P-W, Wang J-Y, Lee H-M (2004) Model selection of SVMs using GA approach. In: 2004 IEEE International Joint Conference on Neural Networks (IEEE Cat. No.04CH37541). pp 2035–2040 vol.3

\bibitem{cochinwala2001}
Cochinwala M, Kurien V, Lalk G, Shasha D (2001) Efficient data reconciliation. Information Sciences 137:1–15. \href{https://doi.org/10.1016/S0020-0255(00)00070-0}{10.1016/S0020-0255(00)00070-0}

\bibitem{cornelis2010}
Cornelis C, Jensen R, Hurtado G, Śle¸zak D (2010) Attribute selection with fuzzy decision reducts. Information Sciences 180:209–224. \href{https://doi.org/10.1016/j.ins.2009.09.008}{ 10.1016/j.ins.2009.09.008}

\bibitem{debaets2018}
De Baets S, Harvey N (2018) Forecasting from time series subject to sporadic perturbations: Effectiveness of different types of forecasting support. International Journal of Forecasting 34:163–180. \href{https://doi.org/10.1016/j.ijforecast.2017.09.007}{ 10.1016/j.ijforecast.2017.09.007}

27.
\bibitem{dey1998}
Dey D, Sarkar S, De P (1998) Entity matching in heterogeneous databases: a distance-based decision model. In: Proceedings of the Thirty-First Hawaii International Conference on System Sciences. pp 305–313 vol.7

\bibitem{droge2012}
Droge C, Vickery SK, Jacobs MA (2012) Does supply chain integration mediate the relationships between product/process strategy and service performance? An empirical study. International Journal of Production Economics 137:250–262. \href{https://doi.org/10.1016/j.ijpe.2012.02.005}{ 10.1016/j.ijpe.2012.02.005}

\bibitem{durbin1975}
Durbin R (1975) Letter: Acid secretion by gastric mucous membrane. American Journal of Physiology-Legacy Content 229:1726–1726. \href{https://doi.org/10.1152/ajplegacy.1975.229.6.1726}{https://doi.org/10.1152/ajplegacy.1975.229.6.1726}
\bibitem{efendigil2009}
Efendigil T, Önüt S, Kahraman C (2009) A decision support system for demand forecasting with artificial neural networks and neuro-fuzzy models: A comparative analysis. Expert Systems with Applications 36:6697–6707. \href{https://doi.org/10.1016/j.eswa.2008.08.058}{ 10.1016/j.eswa.2008.08.058}

\bibitem{elgendy2014}
Elgendy N, Elragal A (2014) Big Data Analytics: A Literature Review Paper. In: Perner P (ed) Advances in Data Mining. Applications and Theoretical Aspects. Springer International Publishing, Cham, pp 214–227

\bibitem{elghazel2015}
Elghazel H, Aussem A (2015) Unsupervised feature selection with ensemble learning. Mach Learn 98:157–180. \href{https://doi.org/10.1007/s10994-013-5337-8}{10.1007/s10994-013-5337-8}

\bibitem{elmagarmid2007}
Elmagarmid AK, Ipeirotis PG, Verykios VS (2007) Duplicate Record Detection: A Survey. IEEE Transactions on Knowledge and Data Engineering 19:1–16. \href{https://doi.org/10.1109/TKDE.2007.250581}{10.1109/TKDE.2007.250581}

\bibitem{escalante2010}
Escalante HJ, Montes M, Sucar E (2010) Ensemble particle swarm model selection. In: The 2010 International Joint Conference on Neural Networks (IJCNN). pp 1–8

\bibitem{estevez2009}
Estevez PA, Tesmer M, Perez CA, Zurada JM (2009) Normalized Mutual Information Feature Selection. IEEE Transactions on Neural Networks 20:189–201. \href{https://doi.org/10.1109/TNN.2008.2005601}{10.1109/TNN.2008.2005601}

\bibitem{farhangfar2007}
Farhangfar A, Kurgan LA, Pedrycz W (2007) A Novel Framework for Imputation of Missing Values in Databases. IEEE Transactions on Systems, Man, and Cybernetics - Part A: Systems and Humans 37:692–709. \href{https://doi.org/10.1109/TSMCA.2007.902631}{10.1109/TSMCA.2007.902631}

\bibitem{fellegi1969}
Fellegi IP, Sunter AB (1969) A Theory for Record Linkage. Journal of the American Statistical Association 64:1183–1210. \href{https://doi.org/10.1080/01621459.1969.10501049}{https://doi.org/10.1080/01621459.1969.10501049}

\bibitem{feurer2019}
Feurer M, Hutter F (2019) Hyperparameter Optimization. In: Hutter F, Kotthoff L, Vanschoren J (eds) Automated Machine Learning: Methods, Systems, Challenges. Springer International Publishing, Cham, pp 3–33


\bibitem{forslund2007}
Forslund H, Jonsson P (2007) The impact of forecast information quality on supply chain performance. International Journal of Operations \& Production Management 27:90–107. \href{https://doi.org/10.1108/01443570710714556}{10.1108/01443570710714556}

\bibitem{friedrichs2005}
Friedrichs F, Igel C (2005) Evolutionary tuning of multiple SVM parameters. Neurocomputing 64:107–117. \href{https://doi.org/10.1016/j.neucom.2004.11.022}{10.1016/j.neucom.2004.11.022}

\bibitem{frohlich2005}
Frohlich H, Zell A (2005) Efficient parameter selection for support vector machines in classification and regression via model-based global optimization. In: Proceedings. 2005 IEEE International Joint Conference on Neural Networks, 2005. pp 1431–1436 vol. 3

\bibitem{gandomi2015}
Gandomi A, Haider M (2015) Beyond the hype: Big data concepts, methods, and analytics. International Journal of Information Management 35:137–144. \href{https://doi.org/10.1016/j.ijinfomgt.2014.10.007}{10.1016/j.ijinfomgt.2014.10.007}

\bibitem{bib2}
García S, Luengo J, Herrera F (2015) Data Preparation Basic Models. In: García S, Luengo J, Herrera F (eds) Data Preprocessing in Data Mining. Springer International Publishing, Cham, pp 39–57

\bibitem{bib3}
 García S, Luengo J, Herrera F (2015) Data Reduction. In: García S, Luengo J, Herrera F (eds) Data Preprocessing in Data Mining. Springer International Publishing, Cham, pp 147–162

\bibitem{bib4}
 García S, Luengo J, Herrera F (2015) Dealing with Missing Values. In: García S, Luengo J, Herrera F (eds) Data Preprocessing in Data Mining. Springer International Publishing, Cham, pp 59–105

\bibitem{bib5}
 García S, Luengo J, Herrera F (2015) Discretization. In: García S, Luengo J, Herrera F (eds) Data Preprocessing in Data Mining. Springer International Publishing, Cham, pp 245–283

\bibitem{garnier2019}
Garnier R, Belletoile A (2019) A multi-series framework for demand forecasts in E-commerce

\bibitem{grewal2017}
Grewal D, Roggeveen AL, Nordfält J (2017) The Future of Retailing. Journal of Retailing 93:1–6. \href{https://doi.org/10.1016/j.jretai.2016.12.008}{10.1016/j.jretai.2016.12.008}

\bibitem{grzymala2005}
Grzymala-Busse JW, Goodwin LK, Grzymala-Busse WJ, Zheng X (2005) Handling Missing Attribute Values in Preterm Birth Data Sets. In: Slezak D, Yao J, Peters JF, et al. (eds) Rough Sets, Fuzzy Sets, Data Mining, and Granular Computing. Springer, Berlin, Heidelberg, pp 342–351

\bibitem{guha2004}
Guha S, Koudas N, Marathe A, Srivastava D (2004) Merging the results of approximate match operations. In: Proceedings of the Thirtieth International Conference on Very Large Data Bases - Volume 30. pp 636–647

\bibitem{gunasekaran2001}
Gunasekaran A, Patel C, Tirtiroglu E (2001) Performance measures and metrics in a supply chain environment. International Journal of Operations \& Production Management 21:71–87. \href{https://doi.org/10.1108/01443570110358468}{10.1108/01443570110358468}

\bibitem{gupta2003}
Gupta A, Maranas CD (2003) Managing demand uncertainty in supply chain planning. Computers \& Chemical Engineering 27:1219–1227. \href{https://doi.org/10.1016/S0098-1354(03)00048-6}{10.1016/S0098-1354(03)00048-6}

\bibitem{gurnani2017}
Gurnani M, Korke Y, Shah P, et al. (2017) Forecasting of sales by using fusion of machine learning techniques. In: 2017 International Conference on Data Management, Analytics and Innovation (ICDMAI). pp 93–101

\bibitem{haider2022}
Haider SA, Sajid M, Sajid H, et al. (2022) Deep learning and statistical methods for short- and long-term solar irradiance forecasting for Islamabad. Renewable Energy 198:51–60. \href{https://doi.org/10.1016/j.renene.2022.07.136}{10.1016/j.renene.2022.07.136}

\bibitem{hamdan2023}
Hamdan IKA, Aziguli W, Zhang D, Sumarliah E (2023) Machine learning in supply chain: prediction of real-time e-order arrivals using ANFIS. Int J  Syst  Assur  Eng  Manag 14:549–568. \href{https://doi.org/10.1007/s13198-022-01851-7}{10.1007/s13198-022-01851-7}

\bibitem{han2012}
Han J, Kamber M, Pei J (2012) 3 - Data Preprocessing. In: Han J, Kamber M, Pei J (eds) Data Mining (Third Edition). Morgan Kaufmann, Boston, pp 83–124

\bibitem{hardgrave2009}
Hardgrave BC, Aloysius J, Goyal S (2009) Does RFID improve inventory accuracy? A preliminary analysis. International Journal of RF Technologies: Research and Applications 1:44–56. \href{https://doi.org/10.1080/17545730802338333}{10.1080/17545730802338333}

\bibitem{hassanzadeh2009}
Hassanzadeh O, Chiang F, Lee HC, Miller RJ (2009) Framework for evaluating clustering algorithms in duplicate detection. Proc VLDB Endow 2:1282–1293. \href{https://doi.org/10.14778/1687627.1687771}{10.14778/1687627.1687771}

\bibitem{hazen2018}
Hazen BT, Skipper JB, Boone CA, Hill RR (2018) Back in business: operations research in support of big data analytics for operations and supply chain management. Ann Oper Res 270:201–211. \href{https://doi.org/10.1007/s10479-016-2226-0}{ 10.1007/s10479-016-2226-0}

\bibitem{hitchcock2023}
Hitchcock C (2023) Causal Models. In: Zalta EN, Nodelman U (eds) The Stanford Encyclopedia of Philosophy, Spring 2023. Metaphysics Research Lab, Stanford University

\bibitem{ho2007}
Ho C-J (2007) Measuring system performance of an ERP-based supply chain. International Journal of Production Research 45:1255–1277. \href{https://doi.org/10.1080/00207540600635235}{ 10.1080/00207540600635235}

\bibitem{holmqvist2006}
Holmqvist M, Stefansson G (2006) ‘Smart Goods’ and Mobile Rfid a Case with Innovation from Volvo. Journal of Business Logistics 27:251–272. \href{https://doi.org/10.1002/j.2158-1592.2006.tb00225.x}{10.1002/j.2158-1592.2006.tb00225.x}

\bibitem{honghai2005}
Honghai F, Guoshun C, Cheng Y, et al. (2005) A SVM Regression Based Approach to Filling in Missing Values. In: Khosla R, Howlett RJ, Jain LC (eds) Knowledge-Based Intelligent Information and Engineering Systems. Springer, Berlin, Heidelberg, pp 581–587

\bibitem{hu2023}
Hu H, Xu J, Liu M, Lim MK (2023) Vaccine supply chain management: An intelligent system utilizing blockchain, IoT and machine learning. Journal of Business Research 156:113480. \href{https://doi.org/10.1016/j.jbusres.2022.113480}{10.1016/j.jbusres.2022.113480}

\bibitem{javed2012}
Javed K, Babri HA, Saeed M (2012) Feature Selection Based on Class-Dependent Densities for High-Dimensional Binary Data. IEEE Transactions on Knowledge and Data Engineering 24:465–477. \href{https://doi.org/10.1109/TKDE.2010.263}{10.1109/TKDE.2010.263}

\bibitem{javeri2021}
Javeri IY, Toutiaee M, Arpinar IB, et al (2021) Improving Neural Networks for Time Series Forecasting using Data Augmentation and AutoML

\bibitem{jensen2007}
Jensen R, Shen Q (2007) Fuzzy-Rough Sets Assisted Attribute Selection. IEEE Transactions on Fuzzy Systems 15:73–89. \href{https://doi.org/10.1109/TFUZZ.2006.889761}{10.1109/TFUZZ.2006.889761}

\bibitem{joachims1999}
Joachims T (1999) Making large-scale svm learning practical. advances in kernel methods-support vector learning. \href{http://svmlight.joachims.org/}{svmlight joachims org/}

\bibitem{john1994}
John GH (1994) Cross-validated C4. 5: Using error estimation for automatic parameter selection. Training 3:

\bibitem{johnson2007}
Johnson R, Wichern D (2007) Matrix algebra and random vectors. Applied Multivariate Statistical Analysis, 6th ed; Pearson: Upper Saddle River, NJ, USA 49–110

\bibitem{kalousis2007}
Kalousis A, Prados J, Hilario M (2007) Stability of feature selection algorithms: a study on high-dimensional spaces. Knowl Inf Syst 12:95–116. \href{https://doi.org/10.1007/s10115-006-0040-8}{10.1007/s10115-006-0040-8}

\bibitem{kang2005}
Kang Y, Gershwin SB (2005) Information inaccuracy in inventory systems: stock loss and stockout. IIE Transactions 37:843–859. \href{https://doi.org/10.1080/07408170590969861}{10.1080/07408170590969861}

\bibitem{karingula2021}
Karingula SR, Ramanan N, Tahmasbi R, et al. (2021) Boosted Embeddings for Time Series Forecasting.

\bibitem{kilimci2019}
Kilimci ZH, Akyuz AO, Uysal M, et al. (2019) An Improved Demand Forecasting Model Using Deep Learning Approach and Proposed Decision Integration Strategy for Supply Chain. Complexity 2019:e9067367. \href{https://doi.org/10.1155/2019/9067367}{10.1155/2019/9067367}

\bibitem{kohavi1995}
Kohavi R, John GH (1995) Automatic Parameter Selection by Minimizing Estimated Error. In: Prieditis A, Russell S (eds) Machine Learning Proceedings 1995. Morgan Kaufmann, San Francisco (CA), pp 304–312

\bibitem{jha2021}
Kumar Jha B, Pande S (2021) Time Series Forecasting Model for Supermarket Sales using FB-Prophet. In: 2021 5th International Conference on Computing Methodologies and CommunicaAtion (ICCMC). pp 547–554

\bibitem{kuo2016}
Kuo RJ, Tseng YS, Chen Z-Y (2016) Integration of fuzzy neural network and artificial immune system-based back-propagation neural network for sales forecasting using qualitative and quantitative data. J Intell Manuf 27:1191–1207. \href{https://doi.org/10.1007/s10845-014-0944-1}{10.1007/s10845-014-0944-1}

\bibitem{kwak2002a}
Kwak N, Choi C-H (2002) Input feature selection by mutual information based on Parzen window. IEEE Transactions on Pattern Analysis and Machine Intelligence 24:1667–1671. \href{https://doi.org/10.1109/TPAMI.2002.1114861}{10.1109/TPAMI.2002.1114861}

\bibitem{kwak2002b}
Kwak N, Choi C-H (2002) Input feature selection for classification problems. IEEE Transactions on Neural Networks 13:143–159. \href{https://doi.org/10.1109/72.977291}{10.1109/72.977291}

\bibitem{lapide1999}
Lapide L (1999) New developments in business forecasting. The Journal of Business Forecasting 17:28–29

\bibitem{li2018}
Li C, Lim A (2018) A greedy aggregation–decomposition method for intermittent demand forecasting in fashion retailing. European Journal of Operational Research 269:860–869. \href{https://doi.org/10.1016/j.ejor.2018.02.029}{10.1016/j.ejor.2018.02.029}

\bibitem{li2023}
Li X, Sengupta T, Si Mohammed K, Jamaani F (2023) Forecasting the lithium mineral resources prices in China: Evidence with Facebook Prophet (Fb-P) and Artificial Neural Networks (ANN) methods. Resources Policy 82:103580. \href{https://doi.org/10.1016/j.resourpol.2023.103580}{10.1016/j.resourpol.2023.103580}

\bibitem{little1987}
Little RJ, Rubin DB (2019) Statistical analysis with missing data. John Wiley \& Sons

\bibitem{liu2002}
Liu H, Hussain F, Tan CL, Dash M (2002) Discretization: An Enabling Technique. Data Mining and Knowledge Discovery 6:393–423. \href{https://doi.org/10.1023/A:1016304305535}{1023/A:1016304305535}

\bibitem{liu2009}
Liu H, Sun J, Liu L, Zhang H (2009) Feature selection with dynamic mutual information. Pattern Recognition 42:1330–1339. \href{https://doi.org/10.1016/j.patcog.2008.10.028}{10.1016/j.patcog.2008.10.028}

\bibitem{locke2006}
Locke EA, Latham GP (2006) New Directions in Goal-Setting Theory. Curr Dir Psychol Sci 15:265–268. \href{https://doi.org/10.1111/j.1467-8721.2006.00449.x}{10.1111/j.1467-8721.2006.00449.x}

\bibitem{lopez2020}
Lopez-Arevalo I, Aldana-Bobadilla E, Molina-Villegas A, et al. (2020) A Memory-Efficient Encoding Method for Processing Mixed-Type Data on Machine Learning. Entropy 22:1391. \href{https://doi.org/10.3390/e22121391}{10.3390/e22121391}

\bibitem{loshchilov2016}
Loshchilov I, Hutter F (2016) CMA-ES for Hyperparameter Optimization of Deep Neural Networks

\bibitem{maccarthy2016}
MacCarthy BL, Blome C, Olhager J, et al. (2016) Supply chain evolution – theory, concepts and science. International Journal of Operations \& Production Management 36:1696–1718. \href{https://doi.org/10.1108/IJOPM-02-2016-0080}{10.1108/IJOPM-02-2016-0080}

\bibitem{mehmood2022}
Mehmood F, Ghani MU, Ghafoor H, et al. (2022) EGD-SNet: A computational search engine for predicting an end-to-end machine learning pipeline for Energy Generation \& Demand Forecasting. Applied Energy 324:119754. \href{https://doi.org/10.1016/j.apenergy.2022.119754}{10.1016/j.apenergy.2022.119754}

\bibitem{michie1995}
Michie D, Spiegelhalter DJ, Taylor CC (1994) Machine learning, neural and statistical classification

\bibitem{migueis2022}
Miguéis VL, Pereira A, Pereira J, Figueira G (2022) Reducing fresh fish waste while ensuring availability: Demand forecast using censored data and machine learning. Journal of Cleaner Production 359:131852. \href{https://doi.org/10.1016/j.jclepro.2022.131852}{10.1016/j.jclepro.2022.131852}

\bibitem{minnich2006}
Minnich D, Maier FH (2006) Supply Chain Responsiveness and Efficiency – Complementing or Contradicting Each Other?

\bibitem{mitra2002}
Mitra A, Jain A, Kishore A, Kumar P (2022) A Comparative Study of Demand Forecasting Models for a Multi-Channel Retail Company: A Novel Hybrid Machine Learning Approach. Oper Res Forum 3:58. \href{https://doi.org/10.1007/s43069-022-00166-4}{10.1007/s43069-022-00166-4}

\bibitem{mitra2022} 
Mitra P, Murthy CA, Pal SK (2002) Unsupervised feature selection using feature similarity. IEEE Transactions on Pattern Analysis and Machine Intelligence 24:301–312. \href{https://doi.org/10.1109/34.990133}{10.1109/34.990133}

\bibitem{modha2003}
Modha DS, Spangler WS (2003) Feature Weighting in k-Means Clustering. Machine Learning 52:217–237. \href{https://doi.org/10.1023/A:1024016609528}{10.1023/A:1024016609528}

\bibitem{monge1996}
Monge AE, Elkan C, others (1996) The field matching problem: algorithms and applications. In: Kdd. pp 267–270

\bibitem{neely1995}
Neely A, Gregory M, Platts K (1995) Performance measurement system design: A literature review and research agenda. International Journal of Operations \& Production Management 15:80–116. \href{https://doi.org/10.1108/01443579510083622}{10.1108/01443579510083622}

\bibitem{nithin2022}
Nithin SSJ, Rajasekar T, Jayanthy S, et al. (2022) Retail Demand Forecasting using CNN-LSTM Model. In: 2022 International Conference on Electronics and Renewable Systems (ICEARS). pp 1751–1756

\bibitem{noack2008}
Noack D, Rose O (2008) A simulation based optimization algorithm for slack reduction and workforce scheduling. In: 2008 Winter Simulation Conference. pp 1989–1994

\bibitem{peng2005}
Peng H, Long F, Ding C (2005) Feature selection based on mutual information criteria of max-dependency, max-relevance, and min-redundancy. IEEE Transactions on Pattern Analysis and Machine Intelligence 27:1226–1238. \href{https://doi.org/10.1109/TPAMI.2005.159}{10.1109/TPAMI.2005.159}

\bibitem{potdar2017}
Potdar K, Pardawala TS, Pai CD (2017) A comparative study of categorical variable encoding techniques for neural network classifiers. International journal of computer applications 175:7–9

\bibitem{falkner2018}
Falkner S, Klein A, Hutter F (2018) BOHB: Robust and Efficient Hyperparameter Optimization at Scale. In: Proceedings of the 35th International Conference on Machine Learning. PMLR, pp 1437–1446

\bibitem{rabbitt2001}
Rabbitt P, Osman P, Moore B, Stollery B (2001) There are stable individual differences in performance variability, both from moment to moment and from day to day. The Quarterly Journal of Experimental Psychology Section A 54:981–1003. \href{https://doi.org/10.1080/713756013}{https://doi.org/10.1080/713756013}

\bibitem{rathipriya2023}
Rathipriya R, Abdul Rahman AA, Dhamodharavadhani S, et al. (2023) Demand forecasting model for time-series pharmaceutical data using shallow and deep neural network model. Neural Comput \& Applic 35:1945–1957. \href{https://doi.org/10.1007/s00521-022-07889-9}{10.1007/s00521-022-07889-9}

\bibitem{ravikumar2012}
Ravikumar P, Cohen W (2012) A Hierarchical Graphical Model for Record Linkage

\bibitem{refaat2010}
Refaat M (2010) Data preparation for data mining using SAS. Elsevier

\bibitem{ribeiro2022}
Ribeiro AMNC, do Carmo PRX, Endo PT, et al. (2022) Short- and Very Short-Term Firm-Level Load Forecasting for Warehouses: A Comparison of Machine Learning and Deep Learning Models. Energies 15:750. \href{https://doi.org/10.3390/en15030750}{10.3390/en15030750}

\bibitem{robnik2003}
Robnik-Šikonja M, Kononenko I (2003) Theoretical and Empirical Analysis of ReliefF and RReliefF. Machine Learning 53:23–69. \href{https://doi.org/10.1023/A:1025667309714}{10.1023/A:1025667309714}

\bibitem{rodriguez2010}
Rodriguez-Lujan I, Huerta R, Elkan C, Cruz CS (2010) Quadratic programming feature selection. The Journal of Machine Learning Research 11:1491–1516

\bibitem{roweis2000}
Roweis ST, Saul LK (2000) Nonlinear Dimensionality Reduction by Locally Linear Embedding. Science 290:2323–2326. \href{https://doi.org/10.1126/science.290.5500.2323}{10.1126/science.290.5500.2323}

\bibitem{russom2011}
Russom P, others (2011) Big data analytics. TDWI best practices report, fourth quarter 19:1–34

\bibitem{saeys2007}
Saeys Y, Inza I, Larrañaga P (2007) A review of feature selection techniques in bioinformatics. Bioinformatics 23:2507–2517. \href{https://doi.org/10.1093/bioinformatics/btm344}{ 10.1093/bioinformatics/btm344}

\bibitem{safarishahrbijari2018}
Safarishahrbijari A (2018) Workforce forecasting models: A systematic review. Journal of Forecasting 37:739–753. \href{https://doi.org/10.1002/for.2541}{10.1002/for.2541}

\bibitem{sanders2016}
Sanders NR (2016) How to Use Big Data to Drive Your Supply Chain. California Management Review 58:26–48. \href{https://doi.org/10.1525/cmr.2016.58.3.26}{10.1525/cmr.2016.58.3.26}

\bibitem{sandhya2022}
Sandhya P, Bandi R, Himabindu DD (2022) Stock Price Prediction using Recurrent Neural Network and LSTM. In: 2022 6th International Conference on Computing Methodologies and Communication (ICCMC). pp 1723–1728

\bibitem{schliephake2009}
Schliephake K, Stevens G, Clay S (2009) Making resources work more efficiently – the importance of supply chain partnerships. Journal of Cleaner Production 17:1257–1263. \href{https://doi.org/10.1016/j.jclepro.2009.03.020}{10.1016/j.jclepro.2009.03.020}

\bibitem{shukla2022}
Shukla S, Pillai VM (2022) Stockout Prediction in Multi Echelon Supply Chain using Machine Learning Algorithms

\bibitem{siami2018}
Siami-Namini S, Tavakoli N, Siami Namin A (2018) A Comparison of ARIMA and LSTM in Forecasting Time Series. In: 2018 17th IEEE International Conference on Machine Learning and Applications (ICMLA). pp 1394–1401

\bibitem{singha2022}
Singha D, Panse C (2022) Application of different Machine Learning models for Supply Chain Demand Forecasting: Comparative Analysis. In: 2022 2nd International Conference on Innovative Practices in Technology and Management (ICIPTM). pp 312–318

\bibitem{singla2004}
Singla P, Domingos P (2004) Multi-relational record linkage. In: Proc. KDD-2004 Workshop Multi-Relational Data Mining. pp 31–48

\bibitem{snoek2012}
Snoek J, Larochelle H, Adams RP (2012) Practical bayesian optimization of machine learning algorithms. Advances in neural information processing systems 25:

\bibitem{spitzer1978}
Spitzer JJ (1978) A Monte Carlo Investigation of the Box-Cox Transformation in Small Samples. Journal of the American Statistical Association 73:488–495. \href{https://doi.org/10.2307/2286587}{10.2307/2286587}

\bibitem{strack2010}
Strack G, Pochet Y (2010) An integrated model for warehouse and inventory planning. European Journal of Operational Research 204:35–50. \href{https://doi.org/10.1016/j.ejor.2009.09.006}{ 10.1016/j.ejor.2009.09.006}

\bibitem{sun2013}
Sun Q, Pfahringer B, Mayo M (2013) Towards a Framework for Designing Full Model Selection and Optimization Systems. In: Zhou Z-H, Roli F, Kittler J (eds) Multiple Classifier Systems. Springer, Berlin, Heidelberg, pp 259–270

\bibitem{sun2010}
Sun Y, Todorovic S, Goodison S (2010) Local-Learning-Based Feature Selection for High-Dimensional Data Analysis. IEEE Transactions on Pattern Analysis and Machine Intelligence 32:1610–1626. \href{https://doi.org/10.1109/TPAMI.2009.190}{10.1109/TPAMI.2009.190}

\bibitem{swiniarski2003}
Swiniarski RW, Skowron A (2003) Rough set methods in feature selection and recognition. Pattern Recognition Letters 24:833–849. \href{https://doi.org/10.1016/S0167-8655(02)00196-4}{10.1016/S0167-8655(02)00196-4}

\bibitem{thatte2013}
Thatte AA, Rao SS, Ragu-Nathan TS (2013) Impact Of SCM Practices Of A Firm On Supply Chain Responsiveness And Competitive Advantage Of A Firm. JABR 29:499–530. \href{https://doi.org/10.19030/jabr.v29i2.7653}{10.19030/jabr.v29i2.7653}

\bibitem{tranfield2003}
Tranfield D, Denyer D, Smart P (2003) Towards a Methodology for Developing Evidence-Informed Management Knowledge by Means of Systematic Review. British Journal of Management 14:207–222. \href{https://doi.org/10.1111/1467-8551.00375}{10.1111/1467-8551.00375}

\bibitem{tuv2009}
Tuv E, Borisov A, Runger G, Torkkola K (2009) Feature selection with ensembles, artificial variables, and redundancy elimination. The Journal of Machine Learning Research 10:1341–1366

\bibitem{berg1999}
van den BERG JP (1999) A literature survey on planning and control of warehousing systems. IIE Transactions 31:751–762. \href{https://doi.org/10.1080/07408179908969874}{10.1080/07408179908969874}

\bibitem{vanpoucke2009}
Vanpoucke E, Boyer KK, Vereecke A (2009) Supply chain information flow strategies: an empirical taxonomy. International Journal of Operations \& Production Management 29:1213–1241. \href{https://doi.org/10.1108/01443570911005974}{10.1108/01443570911005974}

\bibitem{varela2014}
Varela Rozados I, Tjahjono B (2014) Big Data Analytics in Supply Chain Management: Trends and Related Research

\bibitem{verykios2000}
Verykios VS, Elmagarmid AK, Houstis EN (2000) Automating the approximate record-matching process. Information Sciences 126:83–98. \href{https://doi.org/10.1016/S0020-0255(00)00013-X}{https://doi.org/10.1016/S0020-0255(00)00013-X}

\bibitem{vithitsoontorn2022}
Vithitsoontorn C, Chongstitvatana P (2022) Demand Forecasting in Production Planning for Dairy Products Using Machine Learning and Statistical Method. In: 2022 International Electrical Engineering Congress (iEECON). pp 1–4

\bibitem{waller2013}
Waller MA, Fawcett SE (2013) Data Science, Predictive Analytics, and Big Data: A Revolution That Will Transform Supply Chain Design and Management. Journal of Business Logistics 34:77–84. \href{https://doi.org/10.1111/jbl.12010}{10.1111/jbl.12010}

\bibitem{waller2006}
Waller MA, Nachtmann H, Hunter J (2006) Measuring the impact of inaccurate inventory information on a retail outlet. The International Journal of Logistics Management 17:355–376. \href{https://doi.org/10.1108/09574090610717527}{10.1108/09574090610717527}

\bibitem{wan2021}
Wan Y, Chen Y, Yan C, Zhang B (2021) Similarity-based sales forecasting using improved ConvLSTM and prophet. Intelligent Data Analysis 25:383–396. \href{https://doi.org/10.3233/IDA-205103}{https://doi.org/10.3233/IDA-205103}

\bibitem{wang2021}
Wang S, Yang Y (2021) M-GAN-XGBOOST model for sales prediction and precision marketing strategy making of each product in online stores. Data Technologies and Applications 55:749–770. \href{https://doi.org/10.1108/DTA-11-2020-0286}{10.1108/DTA-11-2020-0286}

\bibitem{wang2007}
Wang X, Yang J, Teng X, et al. (2007) Feature selection based on rough sets and particle swarm optimization. Pattern Recognition Letters 28:459–471. \href{https://doi.org/10.1016/j.patrec.2006.09.003}{10.1016/j.patrec.2006.09.003}

\bibitem{wei2007}
Wei H, Billings SA (2007) Feature Subset Selection and Ranking for Data Dimensionality Reduction. IEEE Transactions on Pattern Analysis and Machine Intelligence 29:162–166. \href{https://doi.org/10.1109/TPAMI.2007.250607}{10.1109/TPAMI.2007.250607}

\bibitem{wei2021}
Wei H, Zeng Q (2021) Research on sales Forecast based on XGBoost-LSTM algorithm Model. J Phys: Conf Ser 1754:012191. \href{https://doi.org/10.1088/1742-6596/1754/1/012191}{10.1088/1742-6596/1754/1/012191}

\bibitem{wieland2013}
Wieland A (2013) Selecting the right supply chain based on risks. Journal of Manufacturing Technology Management 24:652–668. \href{https://doi.org/10.1108/17410381311327954}{10.1108/17410381311327954}

\bibitem{wilson2011}
Wilson DR (2011) Beyond probabilistic record linkage: Using neural networks and complex features to improve genealogical record linkage. In: The 2011 International Joint Conference on Neural Networks. pp 9–14

\bibitem{winkler1997}
Winkler WE, others (1993) Improved decision rules in the fellegi-sunter model of record linkage. Bureau of the Census Washington, DC

\bibitem{wu2005}
Wu K (2005) An examination of variability and its basic properties for a factory. IEEE Transactions on Semiconductor Manufacturing 18:214–221. \href{https://doi.org/10.1109/TSM.2004.840525}{10.1109/TSM.2004.840525}

\bibitem{yang2021}
Yang Y, Wu Y, Wang P, Jiali X (2021) Stock Price Prediction Based on XGBoost and LightGBM. E3S Web Conf 275:01040. \href{https://doi.org/10.1051/e3sconf/202127501040}{10.1051/e3sconf/202127501040}

\bibitem{yu2018}
Yu W, Chavez R, Jacobs MA, Feng M (2018) Data-driven supply chain capabilities and performance: A resource-based view. Transportation Research Part E: Logistics and Transportation Review 114:371–385. \href{https://doi.org/10.1016/j.tre.2017.04.002}{10.1016/j.tre.2017.04.002}

\bibitem{yue2007}
Yue L, Yafeng Y, Junjun G, Chongli T (2007) Demand Forecasting by Using Support Vector Machine. In: Third International Conference on Natural Computation (ICNC 2007). pp 272–276. 

\bibitem{yue2010}
Yue L, Zhenjiang L, Yafeng Y, et al. (2010) Selective and Heterogeneous SVM Ensemble for Demand Forecasting. In: 2010 10th IEEE International Conference on Computer and Information Technology. pp 1519–1524.

\bibitem{zhao2013}
Zhao Z, Zhang R, Cox J, et al. (2013) Massively parallel feature selection: an approach based on variance preservation. Mach Learn 92:195–220. \href{https://doi.org/10.1007/s10994-013-5373-4}{10.1007/s10994-013-5373-4}

\bibitem{zohdi2022}
Zohdi M, Rafiee M, Kayvanfar V, Salamiraad A (2022) Demand forecasting based machine learning algorithms on customer information: an applied approach. Int j inf tecnol 14:1937–1947. \href{https://doi.org/10.1007/s41870-022-00875-3}{https://doi.org/10.1007/s41870-022-00875-3}


\end{thebibliography}
\end{document}